\documentclass{article}

\usepackage[final]{neurips_mlsb_2025}
\usepackage{multirow}

\usepackage[utf8]{inputenc} 
\usepackage[T1]{fontenc}    
\usepackage{hyperref}       
\usepackage{url}            
\usepackage{booktabs}       
\usepackage{amsfonts,amsmath}       
\usepackage{nicefrac}       
\usepackage{microtype}      
\usepackage{xcolor}         
\usepackage{mathtools}
\usepackage{wrapfig}
\usepackage{caption}  
\usepackage{tcolorbox}
\usepackage[ruled,vlined]{algorithm2e}
\usepackage{subcaption}
\usepackage{graphicx}
\usepackage{enumitem}
\usepackage{here}
\makeatletter
\newcommand\float@H{\@floatplacement\@fpsaddto{{H}}}
\makeatother

\title{Seek and You Shall Fold}

\author{%
    Nadav Bojan Sellam$^{*\dagger}$, 
    Meital Bojan$^{*\dagger}$,
     \textbf{Paul Schanda}$^{\dagger}$\textbf{,}
  \textbf{Alex Bronstein}$^{\dagger\ddagger}$ \\
  $^{\dagger}$IST Austria \,
  $^{\ddagger}$Technion, Israel \, \\
  $^*$ Equal contribution
}

\begin{document}

\maketitle

\begin{abstract}

Accurate protein structures are essential for understanding biological function, yet incorporating experimental data into protein generative models remains a major challenge.  
Most predictors of experimental observables are non-differentiable, making them incompatible with gradient-based conditional sampling. 
This is especially limiting in nuclear magnetic resonance, where rich data such as chemical shifts are hard to directly integrate into generative modeling.  
We introduce a framework for \emph{non-differentiable guidance} of protein generative models, coupling a continuous diffusion-based generator with \emph{any} black-box objective via a tailored genetic algorithm.  
We demonstrate its effectiveness across three modalities: pairwise distance constraints, nuclear Overhauser effect restraints, and for the first time chemical shifts.  
These results establish chemical shift guided structure generation as feasible, expose key weaknesses in current predictors, and showcase a general strategy for incorporating diverse experimental signals.  
Our work points toward automated, data-conditioned protein modeling beyond the limits of differentiability.

\end{abstract}

\section{Introduction}

Proteins are the "backbone" of all cellular processes, serving as the structural framework, molecular machines, and regulatory signals that sustain life. Knowing the conformations that proteins adapt -- an ensemble of three-dimensional structures -- is crucial to understanding how they function and how they interact with other proteins, as well as for designing drugs that modulate their actions.
Experimental workflows typically begin with measurements and then seek a structure that best explains the data, but this mapping remains a major bottleneck~\citep{burley2021pdb,markley2017future}. 
Machine learning has transformed protein modeling: AlphaFold~\cite{jumper2021highly} delivers highly accurate structure predictions, while generative models such as RFdiffusion~\cite{watson2023rf} enable controllable design, and Boltz~\cite{passaro2025boltz} learns thermodynamically plausible distributions to sample alternative states.
BioEmu~\cite{bioemu2024} further captures near-equilibrium variability from molecular dynamics (MD) simulations. 
Yet none of these models are designed to directly \emph{condition} structure generation on experimental measurements, forcing practitioners to resort to expensive MD refinement to generate states consistent with data.

A common strategy for protein-data conditioned modeling is to guide generative diffusion models using differentiable forward models~\citep{maddipatlainverse,maddipatla2024chroma,fadini2025alphaprior}. 
Although mature, well-validated predictors exist for many experimental observables, most are non-differentiable and therefore cannot be used directly for gradient-based guidance. 
Consequently, current approaches must either approximate these predictors with differentiable surrogates or restrict themselves to modalities with analytic gradients.
While effective when such models exist, this approach has several drawbacks: 
(i) it requires differentiable models for every observable, 
(ii) gradients are often noisy or ill-conditioned, leading to unstable convergence, 
(iii) guidance strength must be tuned under a fixed diffusion schedule, and 
(iv) the process is prone to overfitting to noise and producing off-manifold or broken structures that must be discarded or relaxed. 
These limitations call for alternative guidance methods that remain robust in the presence of non-differentiable predictors and experimental noise.

\textbf{Our contributions.} 
We introduce a general framework for data-conditioned protein generative modeling that overcomes the limitations of gradient-based approaches by performing \emph{non-differentiable guidance}. 
Specifically:
\begin{enumerate}[leftmargin=*, itemsep=0pt, topsep=0.5pt]
    \item \textbf{General non-differentiable guidance:} We introduce a method to couple a continuous protein structure generator with \emph{any} black-box scoring function, guiding the generator’s outputs using a tailored genetic algorithm. This allows seamless integration of mature but non-differentiable experimental predictors.
    \item \textbf{Application to nuclear magnetic resonance (NMR):} We demonstrate the framework on several experimental and synthetic constraints. For pairwise distances and nuclear Overhauser effects (NOEs), our approach yields strong improvements. For chemical shifts, we provide the first proof-of-concept demonstration of direct generative guidance, a long-standing goal in NMR structural biology.

\end{enumerate}

\textbf{Focus on NMR.}
In this work, we focus on NMR spectroscopy, which reports on local structure and dynamics and remains a key tool for protein characterization~\citep{kleckner2010nmr}. 
Chemical shifts are the most widely measured and information-rich NMR observables, yet predictors such as UCBShift~\cite{li2019ucbshift,ptaszek2024ucbshift2} and SHIFTX2~\cite{han2011shiftx2} are non-differentiable, making them incompatible with gradient-based guidance. 
\begin{wrapfigure}{r}{0.5\textwidth}
\vspace{-15pt}
    \centering
    \includegraphics[width=0.48\textwidth]{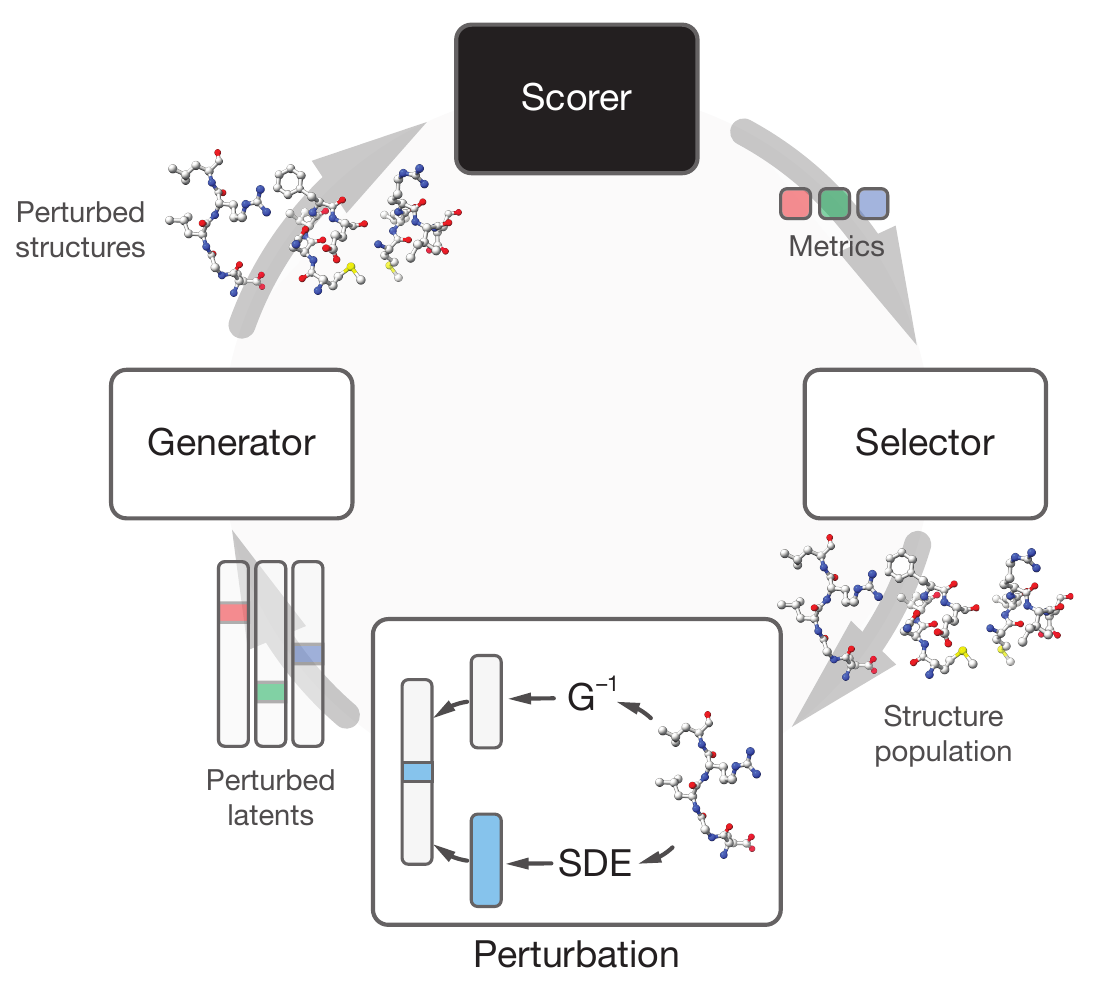}
    \caption{\textbf{Overview of method.} Latents are perturbed and decoded with BioEmu, scored by experimental data, and evolved with a genetic algorithm.}
    \label{fig:alg_flow}
\vspace{-50pt}
\end{wrapfigure}
Integrating chemical shifts into structure generation has long been sought in NMR structural biology but has been infeasible due to this differentiability barrier. 
As chemical shift prediction improves, our framework could dramatically accelerate or even automate peak assignment—one of the most time-consuming bottlenecks in NMR, often requiring days or weeks on high-performance computing resources. 
Additional background is provided in Appendix~\ref{appendix:biological_background}.

\section{Methodology}

We optimize a generator’s latent variables to maximize agreement with experimental data under a black-box score function (Fig.~\ref{fig:alg_flow}). 
Let $x \in \mathcal{X}$ be a protein structure decoded from a latent $z \in \mathcal{Z}$, and let 
$S : \mathcal{X} \to \mathbb{R}$ denote a score derived from experimental observables (e.g., NOEs, chemical shifts). 
Our goal is to find
$
x^\star = \arg\max_{x \in \mathcal{X}} S(x),
$
where $x = G(z_T, 0)$ is obtained by sampling $z_T \sim \mathcal{N}(0,I)$ at diffusion time $T$ and denoising to $t=0$ with the generator $G$. 
Since $S$ is typically non-differentiable, we employ a \emph{protein-specific} genetic algorithm (GA) to search candidate structures, presented in Alg.~\ref{alg:ga}.

\textbf{Generator.}
We use BioEmu~\cite{bioemu2024} as a structural prior.
BioEmu is a lightweight diffusion model, trained to produce thermodynamically consistent protein structures.
Each structure is represented by a residue-wise latent $z \in \mathbb{R}^{N \times d}$, 
with $N$ residues and per-residue dimension $d$, enabling localized perturbations.
We further adapt its stochastic SDE sampler to a deterministic probability flow ODE, making $G(z_T, t)$ reproducible and enabling controlled, repeatable perturbations.

\textbf{Genetic algorithm for structure optimization.}
Our GA~\citep{holland1975adaptation,goldberg1989genetic,eiben2003introduction} maintains a population of protein structures and is designed to keep changes localized while departing from vanilla GAs by:
(1) \emph{disabling crossover}, which would induce destabilizing global jumps;
(2) applying \emph{residue-wise perturbations} in the generator's latent space; and 
(3) introducing a \emph{perturbation time scheduler} that anneals from global to local moves.

\textbf{Residue-wise perturbation:} 
Our perturbation strategy is designed to keep changes localized, preserving stability while still enabling controlled exploration. 
Given a diffusion time $t$, each parent $x^{(i)}$ is mapped to its intermediate latent 
$z_t^{(i)} = G^{-1}(x^{(i)}, t)$ by running the generator backward, so that forwarding $z_t^{(i)}$ to $t{=}0$ exactly reconstructs $x^{(i)}$. 
At the same $t$, a stochastic resample $\hat{z}_t^{(i)}$ is drawn from the reverse SDE, which forwards to a different but distributionally consistent structure.
Next, a residue index $a \sim \mathrm{Uniform}(\{1,\dots,N\})$ is selected, and the corresponding row of $z_t^{(i)}$ is replaced with $\hat{z}_t^{(i)}[a]$, yielding a perturbed latent $z_t^{(i,\mathrm{pert})}$. 
Forwarding $z_t^{(i,\mathrm{pert})}$ to $t{=}0$ then produces the perturbed child structure $x^{(i,\mathrm{child})} = G(z_t^{(i,\mathrm{pert})},0)$.

\textbf{Perturbation time scheduler:} The perturbation time $t_k$ decreases linearly across generations, 
$
    t_k = t_{\max} - \tfrac{k}{K}(t_{\max}-t_{\min}),
$
enabling broad exploration at early steps and fine refinement later. 
This annealing schedule prevents the plateauing observed with a fixed $t$, 
and is directly analogous to annealing molecular dynamics, where the temperature is gradually lowered to transition from broad exploration to local refinement. We provide a more detailed discussion of the scheduler and its impact in Appendix~\ref{appendix:plateau}.


  
  
  
    
    
    
\begin{wrapfigure}{r}{0.57\linewidth}
\vspace{-27pt}
\begin{minipage}{\linewidth}

\begin{algorithm}[H]
\DontPrintSemicolon
\caption{Genetic guidance for protein structures}
\label{alg:ga}

\begin{minipage}{0.97\linewidth} 
\KwIn{scorer $S$; generator $G$; total generations $K$; $t_{\min}, t_{\max}$}

$\{z_T^{(i)}\}_{i=1}^P \sim \mathcal{N}(0,I)$\;

\For{$k=0 \dots K$}{
    \tcp{(1) Scoring}
    $s^{(i)} \gets S(G(z_T^{(i)},0))$\;

    \tcp{(2) Selection}
    Retain elites; sample parents via tournament\;

    \tcp{(3) Controlled perturbation}
    $t_k \gets t_{\max} - \frac{k}{K}(t_{\max}-t_{\min})$\;

    \For{each selected parent $x^{(i)}$}{
        $z_{t_k}^{(i)} \gets G^{-1}(x^{(i)},t_k)$\;
        $\hat z_{t_k}^{(i)} \gets \text{SDE}^{-1}(x^{(i)},t_k)$\;

        $a \sim \mathrm{Uniform}(\{1,\dots,N\})$\;

        $z_{t_k}^{(i)}[a] \gets \hat z_{t_k}^{(i)}[a]$\;

        $x^{(i,\text{child})} \gets G(z_{t_k}^{(i)},0)$\;
    }
    \tcp{(4) Population update}
    Next generation $\gets$ elites $+$ children\;
}

\end{minipage}
\end{algorithm}

\end{minipage}
\vspace{-40pt}
\end{wrapfigure}

This population-based search maintains diversity, avoids greedy local updates that can lead to poor minima, and is not tied to a fixed diffusion schedule, allowing iterative refinement until convergence. 
Additional implementation details are provided in Appendix~\ref{appendix:background} and~\ref{appendix:technical_details}.

\vspace{-7pt}

\section{Results}

\textbf{Experimental setup.} Our experiments test whether genetic guidance can recover accurate folds in proteins where state-of-the-art generators, including BioEmu and AlphaFold3 (AF3), yield erroneous or partially misfolded structures. 
We focus on challenging cases with available NMR data, where both methods struggle to reproduce the correct fold. 
In all experiments, the search was initialized from an AF3 prediction, unless stated otherwise. 
This setting provides a stringent test: success requires not only generating physically plausible structures, but also satisfying experimental constraints that reveal limitations of current predictors. 
We begin with a toy example using pairwise distance restraints, then proceed to experiments with NOE-derived restraints, and finally evaluate chemical shift scoring. 
Complete implementation details are provided in Appendix~\ref{appendix:technical_details}, with additional results in Appendix~\ref{appendix:extended_exp}.

\begin{figure}
\vspace{-10pt}
    \centering
    \includegraphics[width=0.55\linewidth]{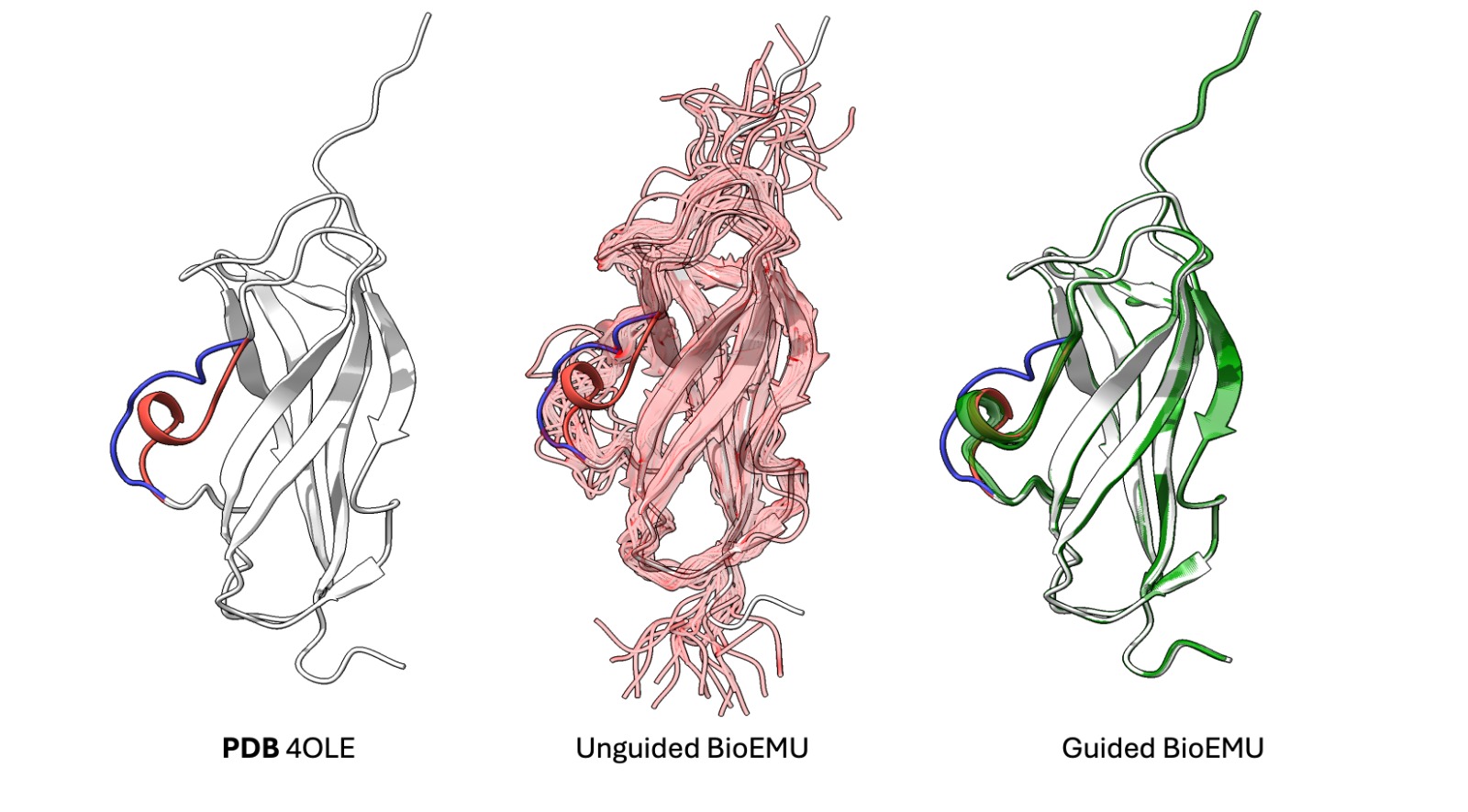}
    \caption{\textbf{Pairwise distance guidance of 4OLE.} 
    Alternative conformations of residues~$60-68$ were guided from conformation~B (blue) toward conformation~A (red) using $C_\alpha$ pairwise distance restraints. 
    Without guidance, BioEmu produces structures resembling conformation~B, whereas guidance enables recovery of the helical conformation~A.}

    \label{fig:4ole_guide}
\end{figure}

\textbf{Pairwise distance guidance}\label{sec:pairwise_results}
As a toy experiment, in preparation for NOE-based restraints, we derived pairwise distance restraints from conformation~A of 4OLE, using $C_\alpha$--$C_\alpha$ distances of at most 5~\AA\ to mimic NOE-derived contacts, with an $L_1$ loss on these distances. This protein exhibits significant alternative conformations in residues~$60$--$68$: conformation~A contains a helical segment, while conformation~B forms a strand. Starting from the conformation~B PDB, we guided the system toward conformation~A. Guidance recovered the helical state, which is otherwise inaccessible to BioEmu alone (Fig.~\ref{fig:4ole_guide}). This shows our method can capture structural differences spanning several residues. The corresponding metric graph is provided in Appendix~\ref{appendix:pairwise_model}.

\textbf{NOE restraints guidance}
We next evaluated guidance using nuclear Overhauser effect (NOE)–derived restraints. NOE cross-peaks arise from through-space dipolar couplings between protons within approximately 5–6~\AA\ and are conventionally converted into distance restraints, making them a key source of spatial proximity information in NMR. As such, they provide a useful test case for assessing the effectiveness of our guidance method in capturing local geometry.  

\begin{figure}[t]
\vspace{-10pt}
    \centering
    \includegraphics[width=0.6\linewidth]{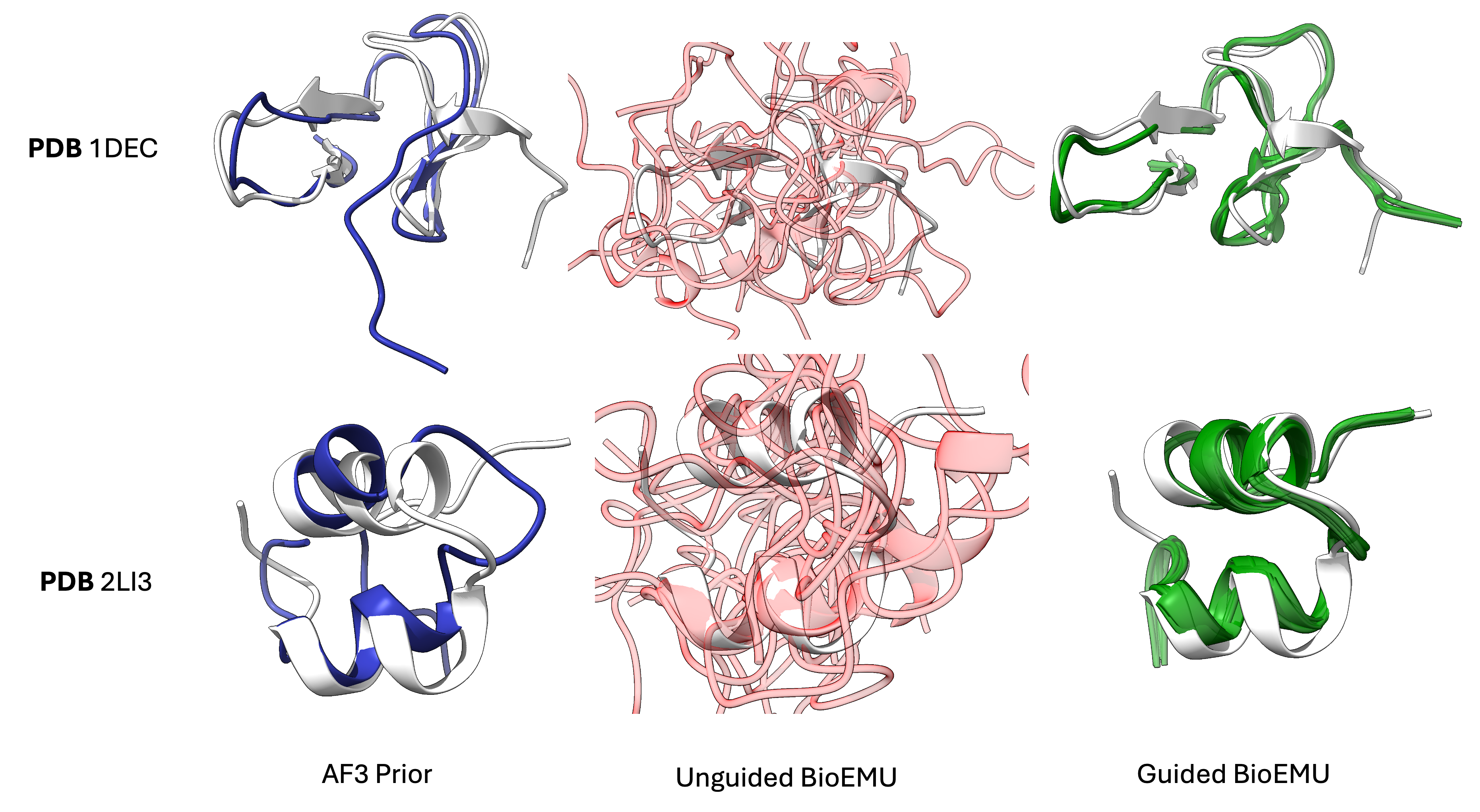}
    \caption{\textbf{NOE-guided structures.} Guided BioEmu with NOE-derived restraints on peptides 1DEC and 2LI3. Reference PDBs are in gray. AF3 priors contained misfolded regions inconsistent with NOEs, unguided BioEmu did not match the PDB, while guided BioEmu reduced violations and recovered the experimental folds.}
    \label{fig:noe_1dec_2li3}
    \vspace{-15pt}
\end{figure}

To assess performance, we used a non-differentiable violation score defined as the fraction of violated restraints multiplied by the average magnitude of their violations (Appendix~\ref{appendix:technical_details}). This metric penalizes both widespread inconsistencies and severe outliers, making it well-suited for guiding structural refinement.
Figure~\ref{fig:noe_1dec_2li3} illustrates results for two peptides (1DEC and 2LI3), with additional cases and quantitative results shown in Appendix~\ref{appendix:additional_noe}. In both cases, AlphaFold predictions served as priors but contained misfolded regions inconsistent with the NOE data. Unguided BioEmu produced incorrect structures, whereas guided BioEmu reduced violations and recovered the experimental folds, showing that NOEs are a strong signal for refinement.

\textbf{Chemical shift guidance}\label{sec:chemical_shift}
\begin{figure}[t]
    \centering
    \begin{subfigure}{0.45\textwidth}
        \centering
        \includegraphics[width=\linewidth]{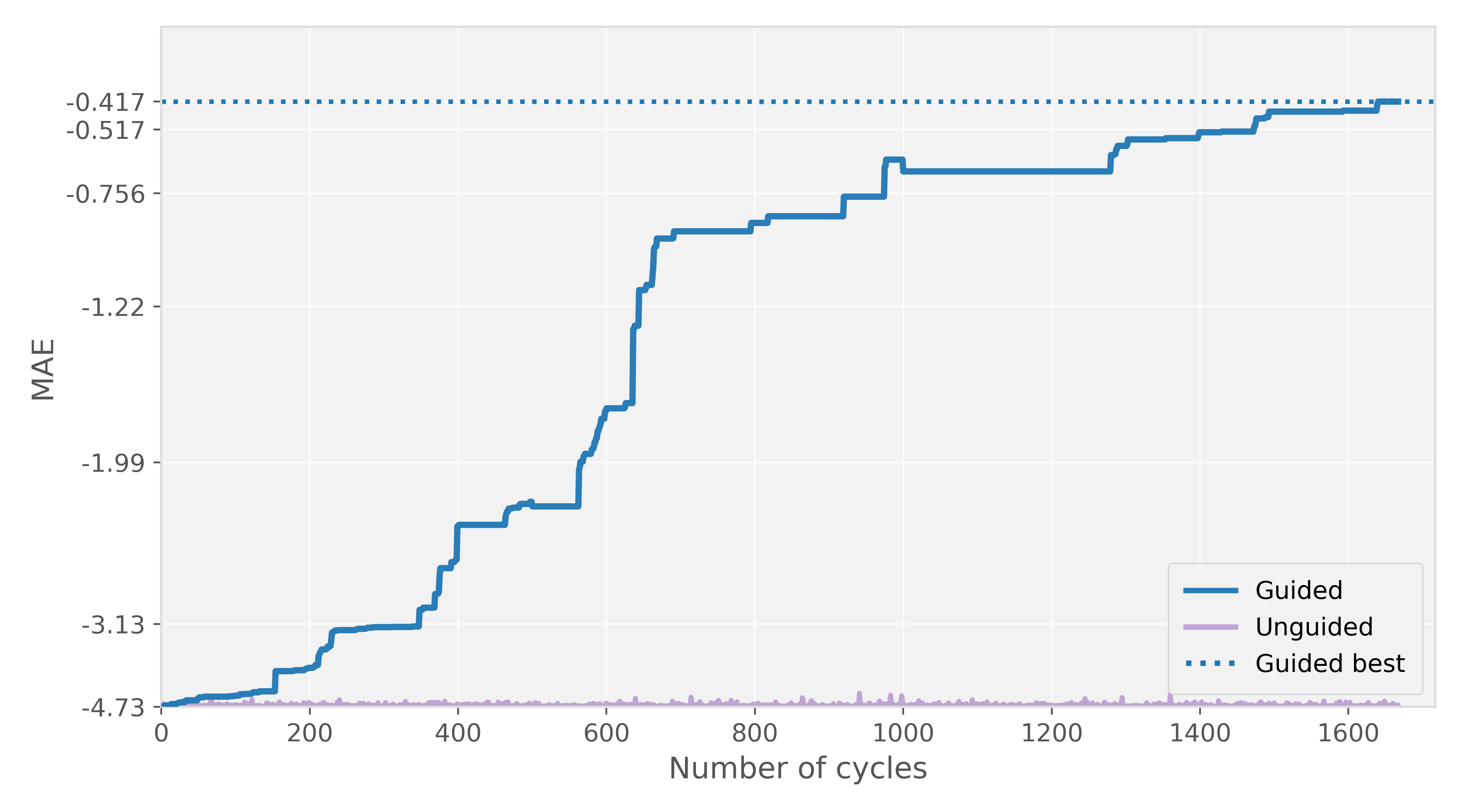}
    \end{subfigure}
    \hfill
    \begin{subfigure}{0.45\textwidth}
        \centering
        \includegraphics[width=\linewidth]{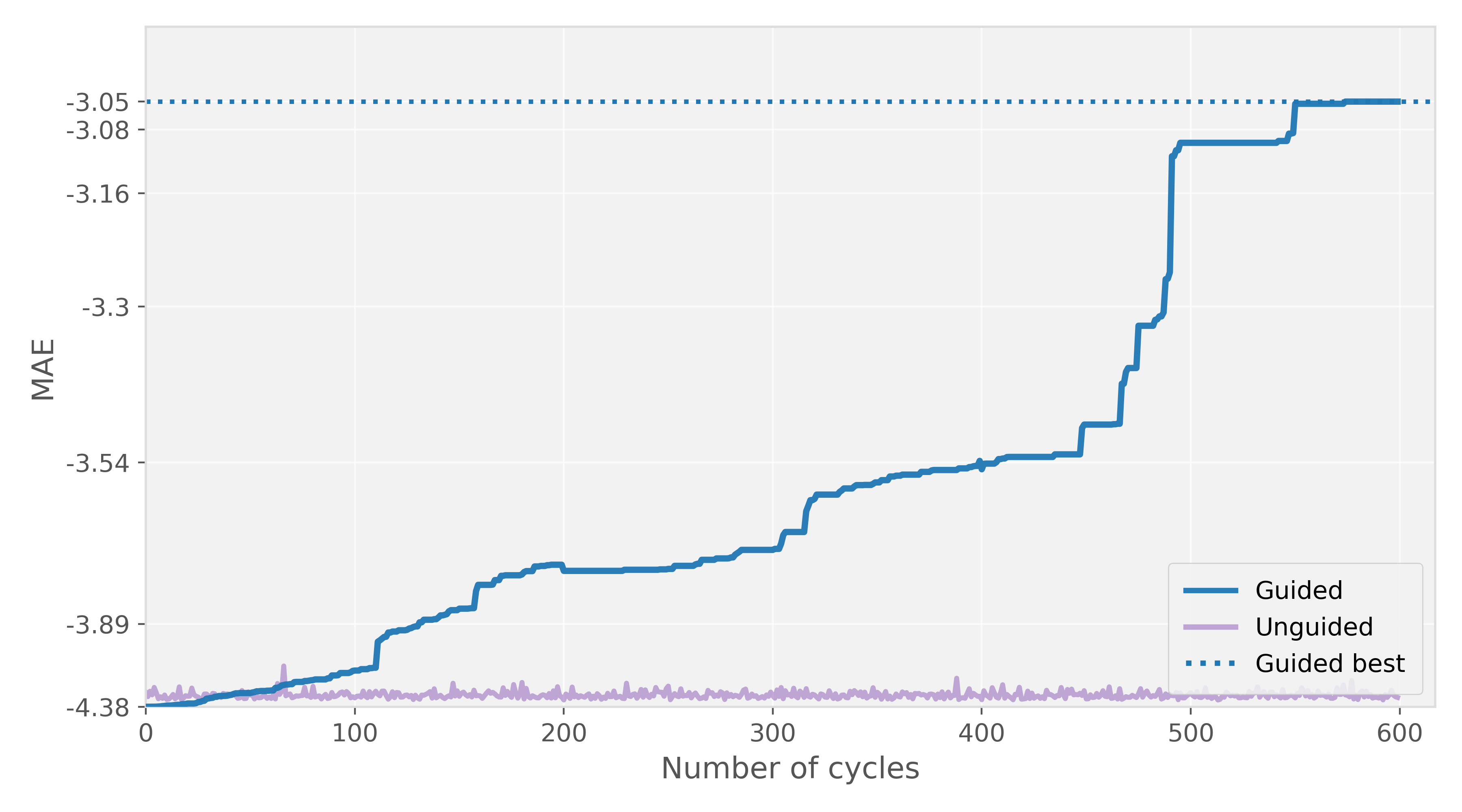}
    \end{subfigure}
    \caption{\textbf{Chemical shift guidance metric.} BioEmu guided using UCBShift-predicted chemical shifts. Plots show mean absolute error (MAE) of the best structure as a function of optimization cycles. \textbf{Left:} peptide \texttt{1DEC} with synthetic PDB-derived shifts. \textbf{Right:} protein \texttt{1DFU} with experimental shifts (BMRB 4395). In both cases, guidance improved the chemical-shift metric, though convergence to the experimental ensemble was not achieved.}
    \label{fig:ucbshift_best_graph}
    \vspace{-15pt}
\end{figure}
The most challenging guidance experiments used chemical shifts, with predicted values for generated structures computed using the widely adopted UCBShift model~\citep{ptaszek2024ucbshift2}. We tested two settings: synthetic shifts for peptide \texttt{1DEC} and experimental shifts for protein \texttt{1DFU}. For \texttt{1DEC} (also used in the NOE experiments), synthetic shifts were generated by running the deposited PDB structure through UCBShift and then attempting to recover them during optimization, providing a controlled baseline fully consistent with the scoring model. For \texttt{1DFU} (BMRB ID~4395), we used calibrated experimental shifts from RefDB~\citep{zhang2003refdb}, selecting this protein because both an NMR ensemble and matched assignments are available.  

In both experiments, guidance was restricted to C$_\alpha$ chemical shifts to emphasize backbone geometry. As shown in Fig.~\ref{fig:ucbshift_best_graph}, the chemical shift objective proved more challenging than pairwise-distance or NOE-based objectives, requiring substantially more optimization cycles to achieve improvement. The score consistently improved over cycles, although plateauing behavior was observed at later stages (Appendix~\ref{appendix:cs_model}). Despite these improvements, the resulting structures did not converge toward the experimental PDB ensemble, indicating that chemical shift agreement does not necessarily translate into structural accuracy. This matches previous reports~\citep{bojan2025representing} that UCBShift predictions provide limited structural signal. Guided structures are provided in Appendix~\ref{appendix:additional_cs}.

\section{Related Work}

Gradient-free and evolutionary approaches are increasingly explored for molecular and biomolecular design. 
Early work such as MUDM~\citep{han2023training} introduced a multi-objective diffusion framework for 3D molecule generation, using posterior approximation and Monte Carlo sampling to compute low-variance conditional gradients without retraining. 
In the protein domain, evolutionary diffusion has been applied to sequence generation in discrete space~\citep{alamdari2023protein}, but this does not address continuous 3D structure generation or conditioning structural samples on experimental data. 
More recently, there has been a surge of concurrent work extending these ideas to new molecular settings, including evolutionary multi-objective diffusion for molecular design~\citep{sun2025evolutionary} and reward-guided fine-tuning via iterative distillation to improve sample efficiency for biomolecular objectives~\citep{su2025iterative}. 
Additional discussion of concurrent work in vision and general diffusion guidance is provided in Appendix~\ref{appendix:related_work}.

While these advances highlight the promise of gradient-free methods, none address the unique challenges of protein structure generation. 
Proteins inhabit a highly constrained, high-dimensional conformational space where even small perturbations can drastically change the resulting structure, and naive search often fails to converge. 
Unlike the discrete sequence domain, where every amino acid sequence corresponds to a valid backbone, the continuous structural domain is far more delicate:  uninformed perturbations in latent space easily push candidates off-manifold, yielding non-physical conformations. 
To our knowledge, this work is the first to develop a tailored evolutionary search method for 3D protein structure generation with respect to \emph{arbitrary} black-box objectives.

\section{Discussion}

We introduced a novel framework for experiment-guided protein structure generation that integrates non-differentiable objectives with generative diffusion models. By coupling BioEmu with a tailored genetic algorithm, we showed that black-box predictors ranging from pairwise distance restraints to NOE violation metrics and chemical shift scores can guide the search toward experimental consistency. Signal strength varied: pairwise restraints served as a proof of concept, NOEs enabled robust recovery of experimental folds, and chemical shifts improved agreement with the scoring model but did not yield structurally accurate ensembles. Limitations arise from both the generator and the guidance signals: BioEmu constrains the accessible conformational space, and current chemical shift predictors such as UCBShift remain limited by insufficient physical grounding, underscoring the need for more accurate and biophysically informed models. Future work includes stronger structural priors, physically informed and differentiable chemical shift models, ensemble-level objectives, and extending the framework to other modalities such as cryo-EM, crosslinking, or spectroscopy. Together, these directions would enable a more faithful integration of experimental data and reinforce genetic guidance as a general strategy for conditioning protein generative models on black-box objectives that fall outside the scope of differentiability.

\begin{ack}
Supported by ISTA and Technion. Thanks to the Bronstein and Schanda labs.
\end{ack}

\newpage


\appendix

\section{Extended Background} \label{appendix:background}

\subsection{BioEmu Training and Sampling}

BioEmu~\cite{bioemu2024} is an E(3)-equivariant conditional diffusion model trained to sample residue-level protein conformations from a learned equilibrium distribution. 
Each protein is represented in a residue-wise backbone frame representation, where residue $i$ is parameterized by a tuple $(r_i, Q_i)$ consisting of the C$_\alpha$ position $r_i \in \mathbb{R}^3$ and a local orientation frame $Q_i \in SO(3)$ constructed via Gram--Schmidt orthogonalization of the N--C$_\alpha$ and C--C$_\alpha$ bonds.

The forward noising process is modeled as a pair of independent stochastic differential equations (SDEs) acting on positions and orientations:
\begin{align}
    d r_t &= -\frac{1}{2} \beta(t) r_t \, dt + \sqrt{\beta(t)} \, dW_t, \\
    d Q_t &= \mathcal{L}_{SO(3)}(Q_t) \, dt + \sqrt{2}\, d\mathcal{W}_t^{SO(3)},
\end{align}
where $\beta(t)$ follows a cosine noise schedule~\cite{nichol2021improved}, $W_t$ is standard Brownian motion in $\mathbb{R}^3$, and $d\mathcal{W}_t^{SO(3)}$ denotes Brownian motion on the rotation group $SO(3)$ with generator $\mathcal{L}_{SO(3)}$ corresponding to the isotropic Gaussian distribution on rotations (IGSO(3))~\cite{de2022riemannian}. 
This formulation ensures that positional noise is variance-preserving and that rotational noise is uniformly distributed over the manifold, enabling the model to learn orientation-equivariant denoising updates.

BioEmu learns a time-dependent score network $s_\theta(r_t, Q_t, t)$ that estimates the gradient of the log-density of the perturbed data:
\begin{equation}
    s_\theta(r_t, Q_t, t) \approx \nabla_{r_t, Q_t} \log p_t(r_t, Q_t),
\end{equation}
and generates samples by integrating the reverse-time SDE with learned score estimates:
\begin{equation}
    d(r_t, Q_t) = \Bigl[-\frac{1}{2}\beta(t) (r_t, Q_t) - \beta(t) s_\theta(r_t, Q_t, t)\Bigr] dt + \sqrt{\beta(t)}\, d\bar{W}_t,
\end{equation}
where $d\bar{W}_t$ denotes time-reversed Brownian motion.

\paragraph{Training data.}  
BioEmu is first pre-trained on clustered AlphaFoldDB structures to encourage conformational diversity. 
It is then fine-tuned on more than 200 ms of equilibrium MD trajectories, reweighted by Markov state models to better match equilibrium populations. 
Finally, it is refined using Property Prediction Fine-Tuning (PPFT) with MEGAscale protein stability measurements, aligning the learned distribution with experimental free energies.

\paragraph{Side-chain modeling.}  
In its original form, BioEmu employs Hpacker~\cite{visani2024h} for reconstructing side-chain conformations from the generated backbones.  
In this work, we instead integrate FlowPacker~\cite{lee2025flowpacker}, a flow-based packing model that provides faster inference and improved physical plausibility of side-chain placements.  
This substitution improves side-chain accuracy while reducing runtime, leading to more reliable structures for downstream genetic guidance.

\subsubsection{DPM-Solver for Fast Sampling}
To generate samples efficiently, BioEmu employs DPM-Solver~\cite{lu2022dpm}, a high-order numerical ODE solver specifically designed for diffusion probabilistic models. 
Given a learned score function $s_\theta(x_t, t)$, the corresponding probability flow ODE is
\begin{equation}
    \frac{d x_t}{dt} = f_\theta(x_t, t) 
    = -\frac{1}{2}\beta(t) \left[x_t + s_\theta(x_t, t)\right],
    \label{eq:probflow}
\end{equation}
where $\beta(t)$ is the noise schedule. 
DPM-Solver integrates this ODE using a second- or third-order solver, yielding high-fidelity samples with significantly fewer function evaluations compared to standard ancestral samplers:
\[
    x_{t_{k-1}} = x_{t_k} + h_k \, \Phi^{(m)}\!\bigl(f_\theta, x_{t_k}, t_k\bigr),
\]
where $h_k = t_{k-1} - t_k$ is the integration step and $\Phi^{(m)}$ is an $m$-th order multi-step update (typically $m=2$ or $m=3$). 
This formulation enables BioEmu to generate thousands of near-equilibrium conformations in only 30--50 denoising steps, compared to hundreds of steps required by Euler--Maruyama or predictor-corrector methods.

DPM-Solver supports both a deterministic ODE solver and a stochastic variant based on the reverse SDE:
\begin{equation}
    d x_t = f_\theta(x_t, t)\, dt + \sqrt{\beta(t)} \, d\bar{W}_t,
    \label{eq:stochastic}
\end{equation}
where $d\bar{W}_t$ is the time-reversed Brownian motion term, reintroducing noise at each step to maintain diversity.
BioEmu's default procedure uses the stochastic variant \eqref{eq:stochastic}, whereas in this work we employ the deterministic ODE formulation \eqref{eq:probflow} to improve prevent drastic changes tot he protein and ensure stable convergence under genetic algorithm–based optimization.

\section{Biological Background}
\label{appendix:biological_background}

\paragraph{NMR spectroscopy for proteins.}
Nuclear magnetic resonance (NMR) spectroscopy is a widely used technique for probing protein structure and dynamics in solution \citep{wuthrich1986nmr}. 
NMR exploits the magnetic properties of atomic nuclei (most commonly $^{1}$H, $^{13}$C, and $^{15}$N) and measures their resonance frequencies in a strong magnetic field. 
Each experiment produces a spectrum in which peaks correspond to specific nuclear spins, providing atom-level information that reflects the local chemical environment and, indirectly, the 3D structure of the protein. 
Unlike X-ray crystallography, NMR is performed in solution, making it particularly valuable for studying flexible regions, conformational exchange, and dynamics \citep{palmer2001nmr}.

\paragraph{Chemical shifts and peak assignment.}
The chemical shift is the fundamental NMR observable: the resonance frequency of a nucleus relative to a reference compound. 
It is highly sensitive to the local electronic environment and therefore encodes detailed information about backbone torsion angles, hydrogen bonding, and secondary structure \citep{wishart2005nmr}. 
In practice, chemical shifts appear as peaks in 1D or 2D spectra, but before they can be used for structure modeling, these peaks must be \emph{assigned} to specific atoms in the protein sequence. 
Peak assignment is notoriously challenging: large proteins produce thousands of overlapping peaks, noise and artifacts can obscure weaker signals, and experimental conditions can shift peaks, making manual assignment a laborious, error-prone process that often requires days or weeks of expert effort \citep{markley2017future}. 
Automating or accelerating chemical shift based assignment remains one of the major bottlenecks in NMR-driven structure determination.

\paragraph{NOE distance restraints.}
Another major source of structural information in NMR is the nuclear Overhauser effect (NOE), which reports on through-space dipolar couplings between nearby protons. 
NOEs are typically measured using nuclear Overhauser effect spectroscopy (NOESY), a 2D or 3D NMR experiment in which cross-peaks indicate spatial proximity between proton pairs. 
These cross-peaks can be converted into approximate upper bounds on inter-proton distances, typically up to 5~Å \citep{wuthrich1986nmr,clore1994nmr}. 
Such distance restraints are invaluable for constraining tertiary structure, as they directly encode spatial proximity between residues. 
However, NOESY spectra are often highly crowded, and many peaks can correspond to multiple possible proton pairs, making NOE assignment a combinatorial problem that usually requires iterative assignment and structure refinement cycles \citep{nilges1997automated}.

\section{Extended Related Work}
\label{appendix:related_work}

Beyond biology, there has been a recent surge of interest in combining diffusion models with evolutionary or black-box search methods in other domains. 
Concurrent work in the vision domain has explored several complementary approaches for gradient-free guidance: large-scale test-time evolutionary search to optimize image and video fidelity~\citep{he2025scaling}, evolutionary algorithms for inference-time alignment of generated samples with downstream objectives~\citep{jajal2025inference}, and black-box conditional generation via evolvable diffusion, which frames conditioning as an optimization problem over the diffusion process~\citep{wei2025evolvable}. 
Dynamic search methods have also been proposed to improve inference-time alignment of diffusion models, offering a gradient-free alternative to direct score-based guidance~\citep{li2025dynamic}.

\section{Extended experiments}\label{appendix:extended_exp}

\subsection{Mean vs.\ best metric}
\label{appendix:mean_vs_best}

Throughout the paper, unless otherwise noted, we report the best-structure metric per cycle, defined as the minimum loss (or maximum score) observed within the population at that cycle. This choice provides a consistent summary across modalities and emphasizes the emergence of the highest-quality candidates during the search.

\begin{figure}[H]
    \centering
    \begin{subfigure}{0.32\textwidth}
        \centering
        \includegraphics[width=\linewidth]{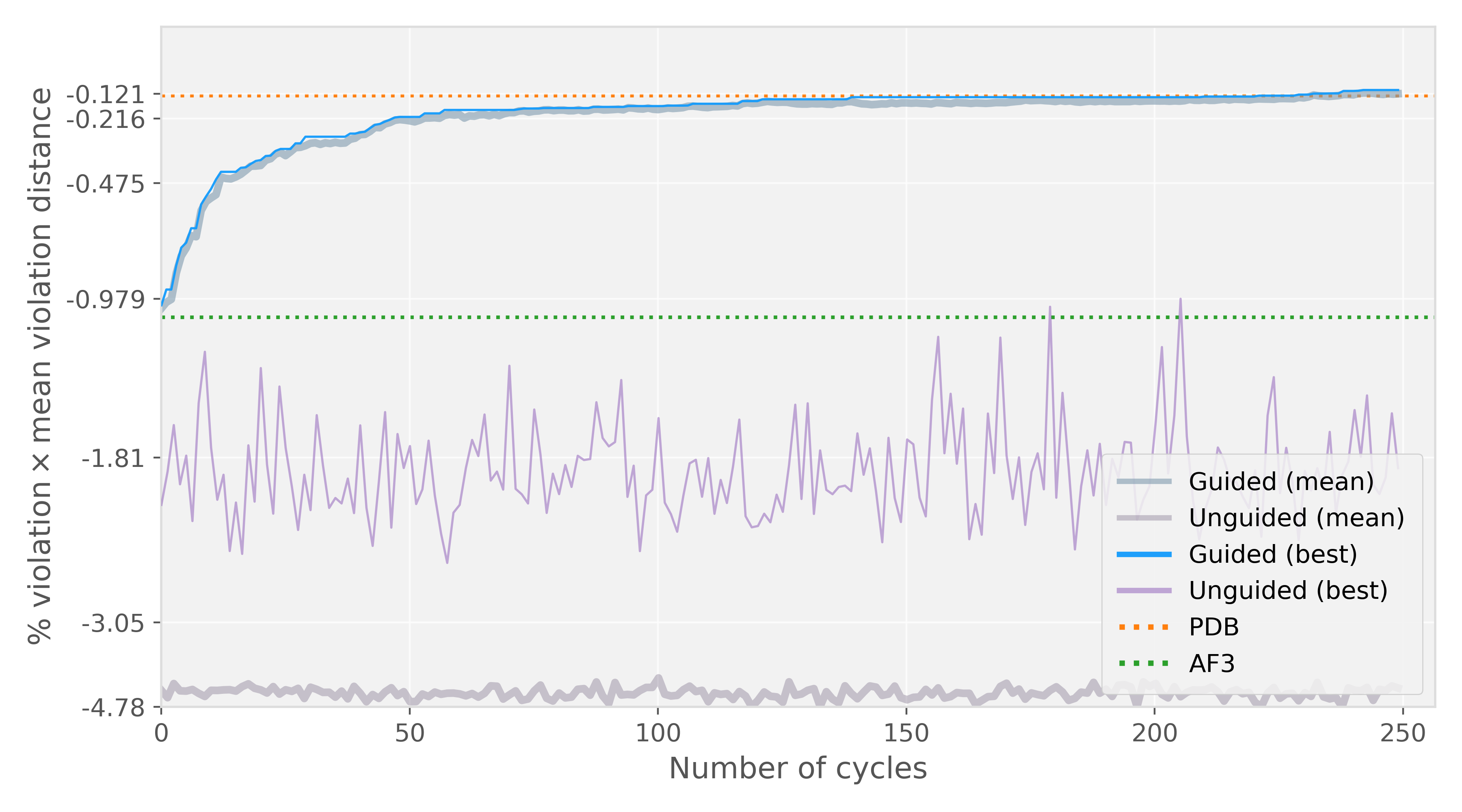}
        \caption{1DEC}
    \end{subfigure}\hfill
    \begin{subfigure}{0.32\textwidth}
        \centering
        \includegraphics[width=\linewidth]{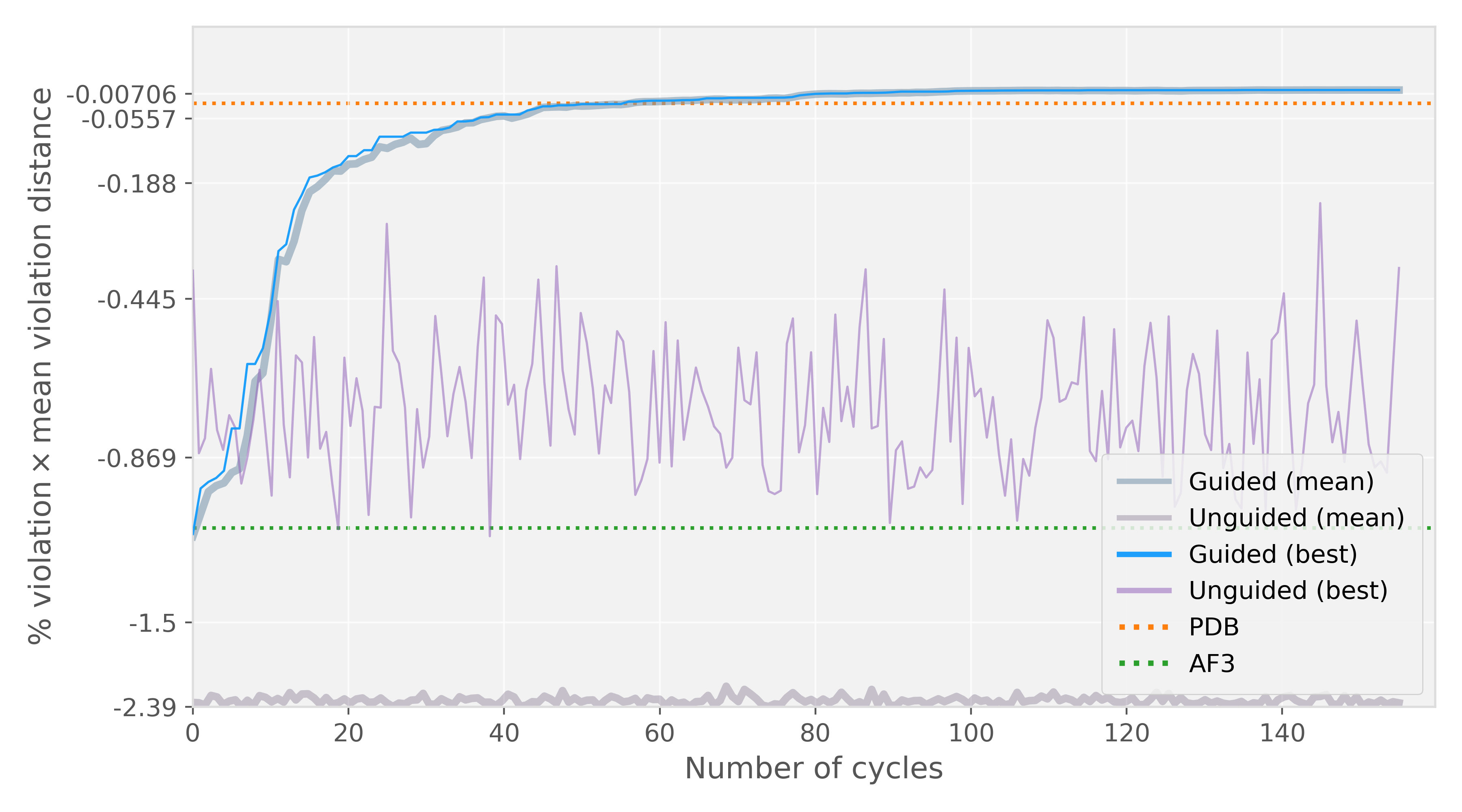}
        \caption{2LI3}
    \end{subfigure}\hfill
    \begin{subfigure}{0.32\textwidth}
        \centering
        \includegraphics[width=\linewidth]{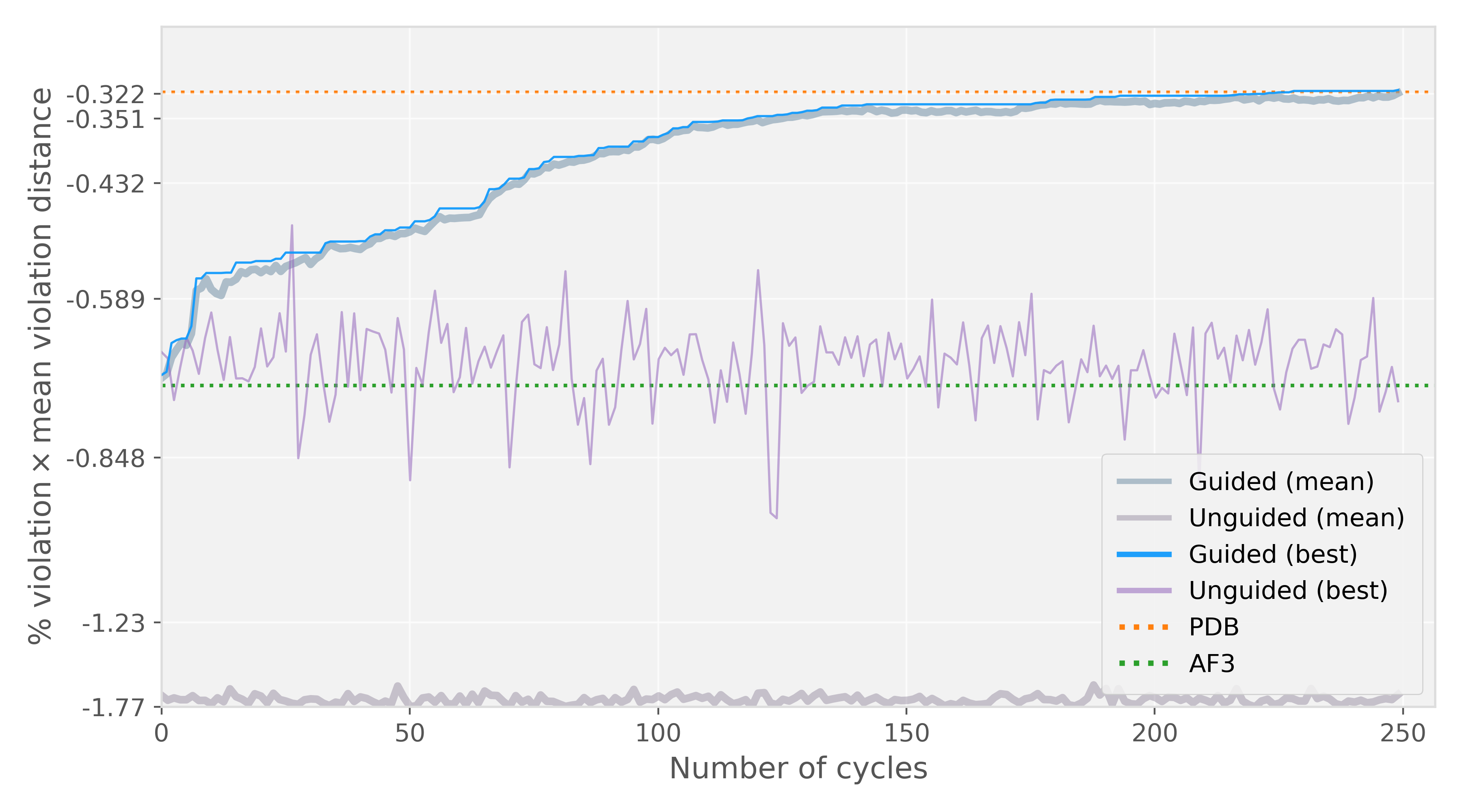}
        \caption{3BBG}
    \end{subfigure}
    \caption{\textbf{Best vs.\ mean metric per cycle.}
    For each target, we plot the best-structure metric (solid) and the mean-structure metric (faint) using the same definitions as in the main text for NOE guidance (Y-axis: fraction of violated restraints $\times$ mean violation distance; X-axis: number of cycles). Dashed baselines indicate PDB and AF3.}
    \label{fig:mean_vs_best_metric}
\end{figure}

For completeness, Fig.~\ref{fig:mean_vs_best_metric} shows the mean-structure metric per cycle plotted alongside the best-structure metric using the same definitions as in the main text. Across targets and modalities, the mean traces closely mirror the best traces and exhibit the same trends (improvements and occasional plateaus), supporting our choice to emphasize the best-structure metric in the main figures.

\subsection{Pairwise distances experiments}
\label{appendix:additional_pairwise}

\begin{figure}[H]
    \centering
    \includegraphics[width=0.7\linewidth]{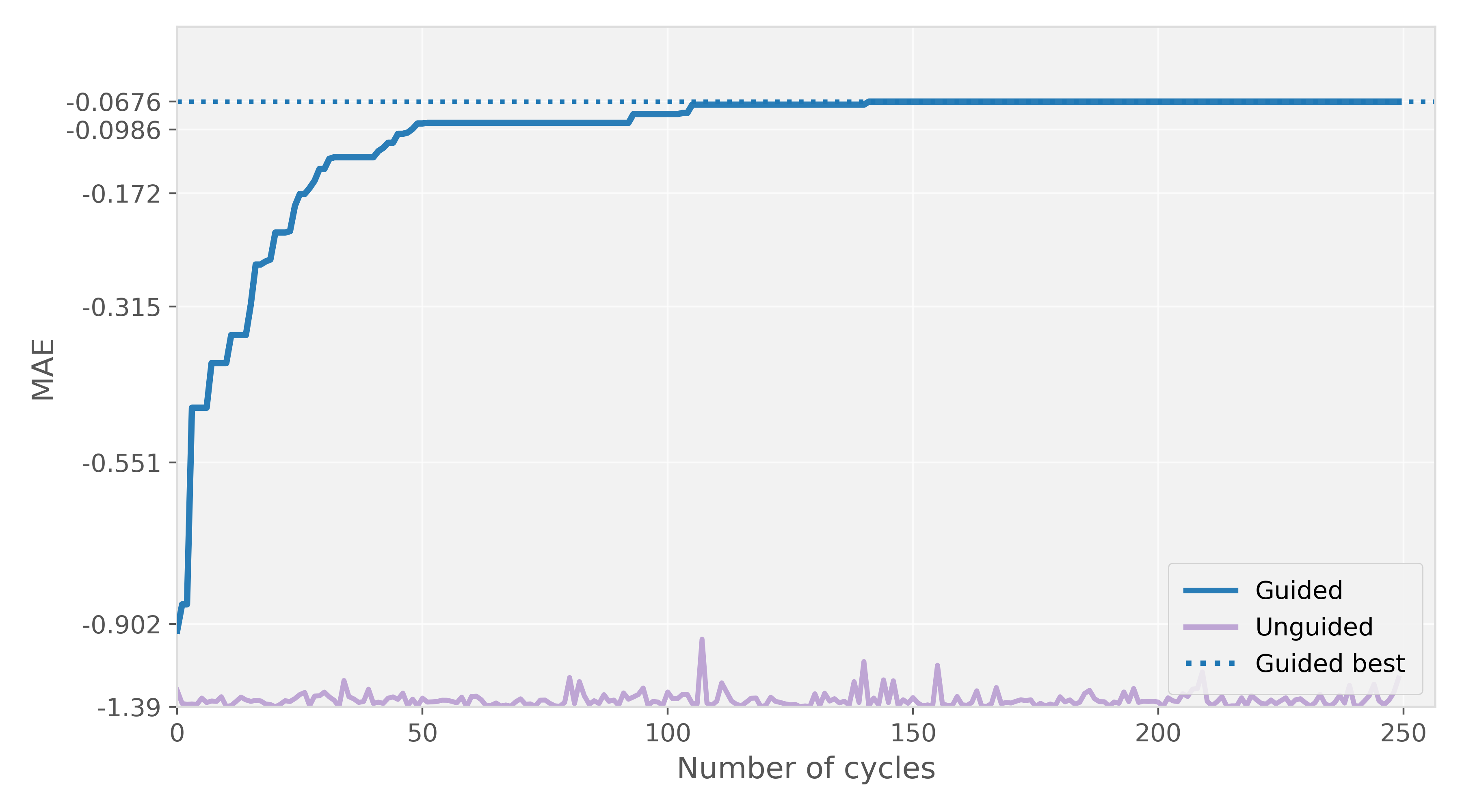}
     \caption{\textbf{4OLE pairwise distance guidance.} 
    BioEmu guided with $C_\alpha$ pairwise distance restraints. 
    Y-axis: mean absolute error (MAE). 
    The search was initialized from altloc B, which BioEmu alone resembles but never recovers as conformation A. 
    Guidance enables consistent improvement in the best-structure metric and recovery of the helical conformation A.}
    \label{fig:pairwise_metric}
\end{figure}

This section provides extended results for pairwise distance guidance.  
As described in Section~\ref{sec:pairwise_results}, pairwise $C_\alpha$--$C_\alpha$ distances within 5~\AA\ were used as surrogate NOE restraints to guide the search.  
Figure~\ref{fig:pairwise_metric} reports results for 4OLE, which exhibits two alternate conformations (A: helical, B: strand).  
Unguided BioEmu produces structures resembling altloc B but never recovers conformation A, whereas guidance consistently improves the best-structure metric and enables recovery of the helical conformation.  
These results illustrate how pairwise distance information can provide a strong corrective signal, even in challenging cases where the unguided prior fails.

\subsection{NOE experiments}
\label{appendix:additional_noe}

\begin{figure}[H]
    \centering
    \includegraphics[width=0.7\linewidth]{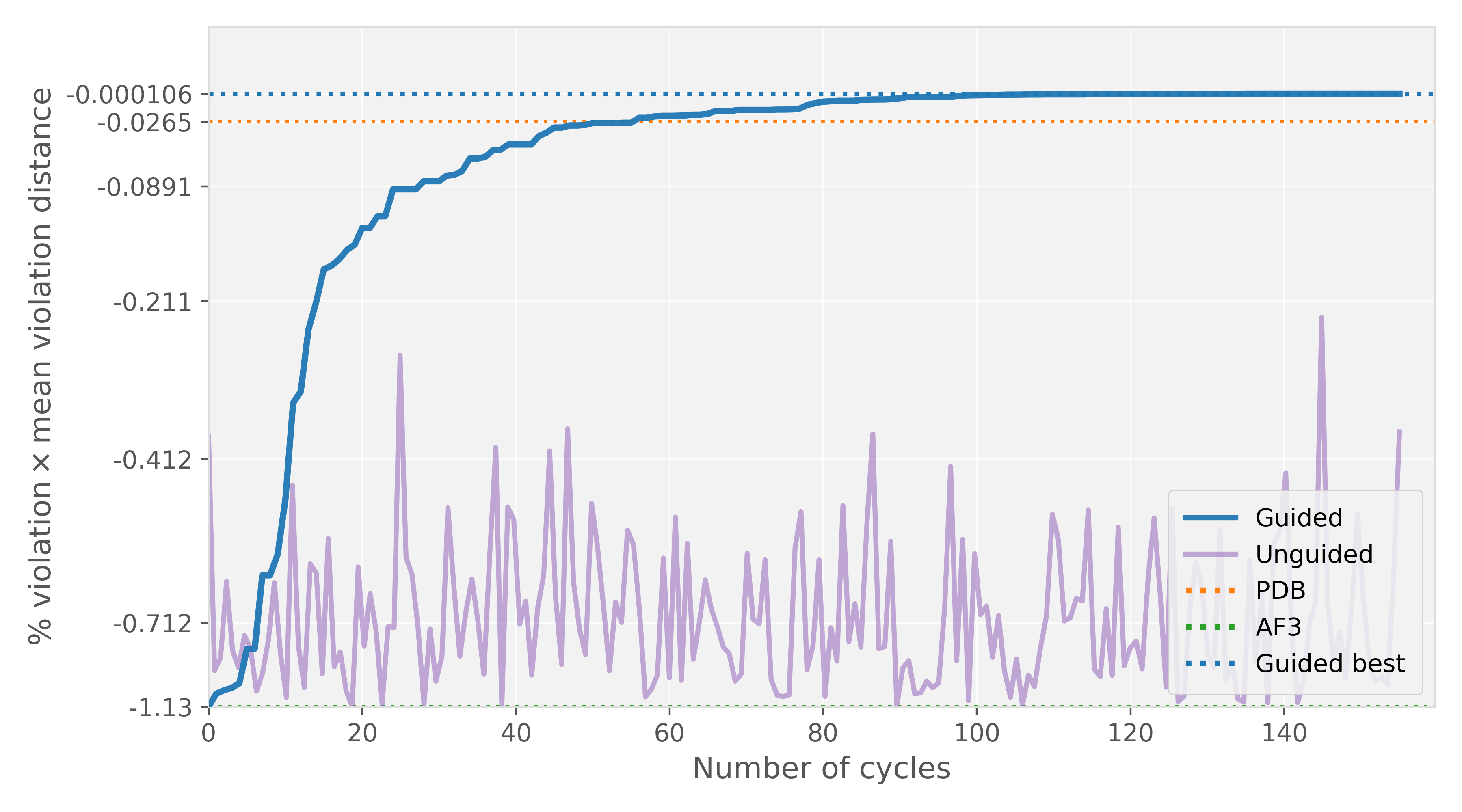}
     \caption{\textbf{2LI3 NOE guidance.} 
    BioEmu guided with NOE-derived restraints. 
    Y-axis: fraction of violated restraints multiplied by mean violation distance. 
    Guidance yields consistent improvement in the best-structure metric, surpassing the PDB according to this measure.}
    \label{fig:noe_2li3}
\end{figure}

\begin{figure}[H]
    \centering
    \includegraphics[width=0.7\linewidth]{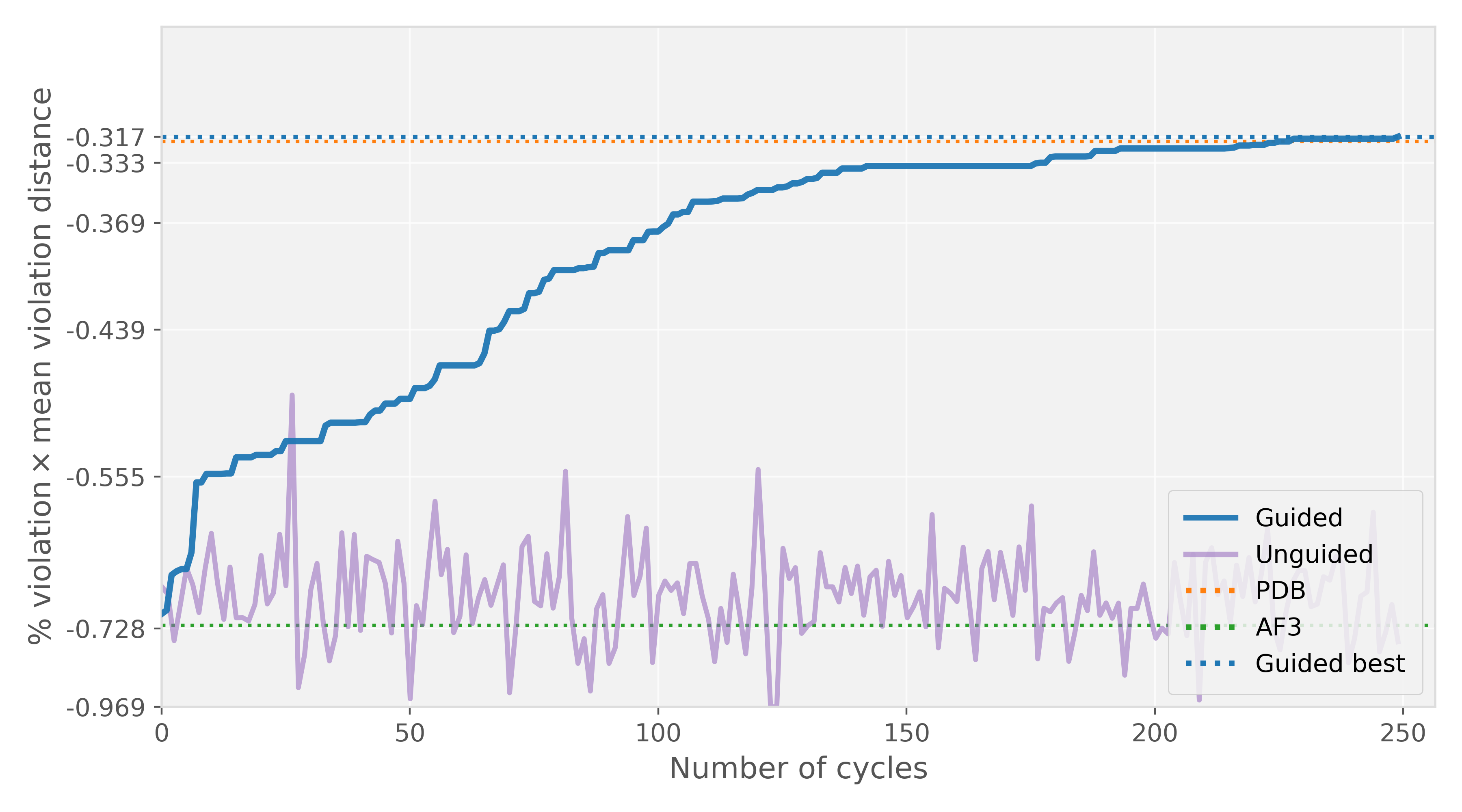}
     \caption{\textbf{3BBG NOE guidance.} 
BioEmu guided with NOE-derived restraints. 
Y-axis: fraction of violated restraints multiplied by mean violation distance. 
Guidance yields consistent improvement in the best-structure metric, reaching performance comparable to the PDB according to this measure.}
    \label{fig:noe_3bbg}
\end{figure}

\begin{figure}[H]
    \centering
    \includegraphics[width=0.7\linewidth]{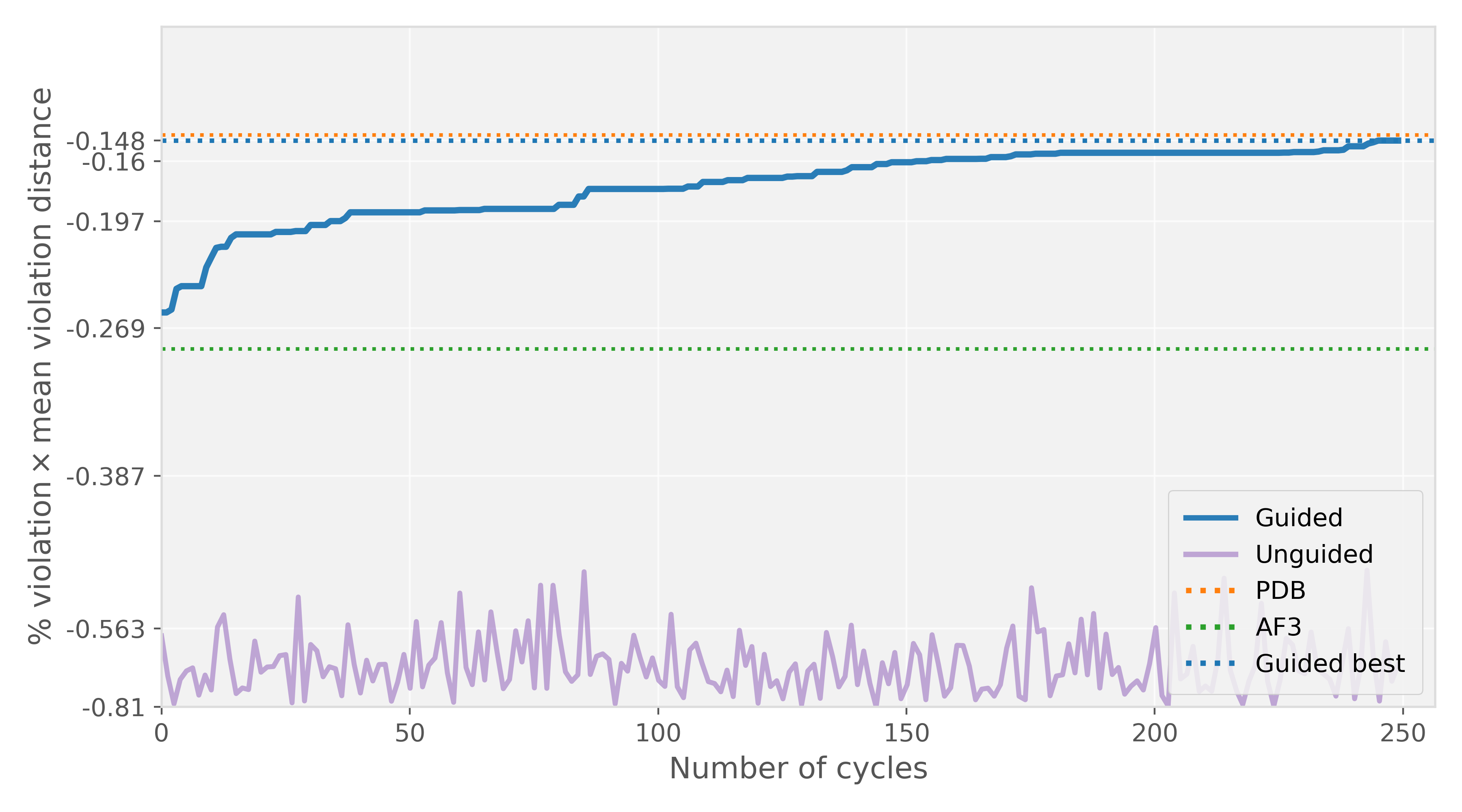}
     \caption{\textbf{2KKZ NOE guidance.} 
    BioEmu guided with NOE-derived restraints. 
    Y-axis: fraction of violated restraints multiplied by mean violation distance. 
    Guidance yields consistent improvement in the best-structure metric, reaching performance comparable to the PDB according to this measure.}
    \label{fig:noe_2kkz}
\end{figure}

\begin{figure}[H]
    \centering
    \includegraphics[width=0.7\linewidth]{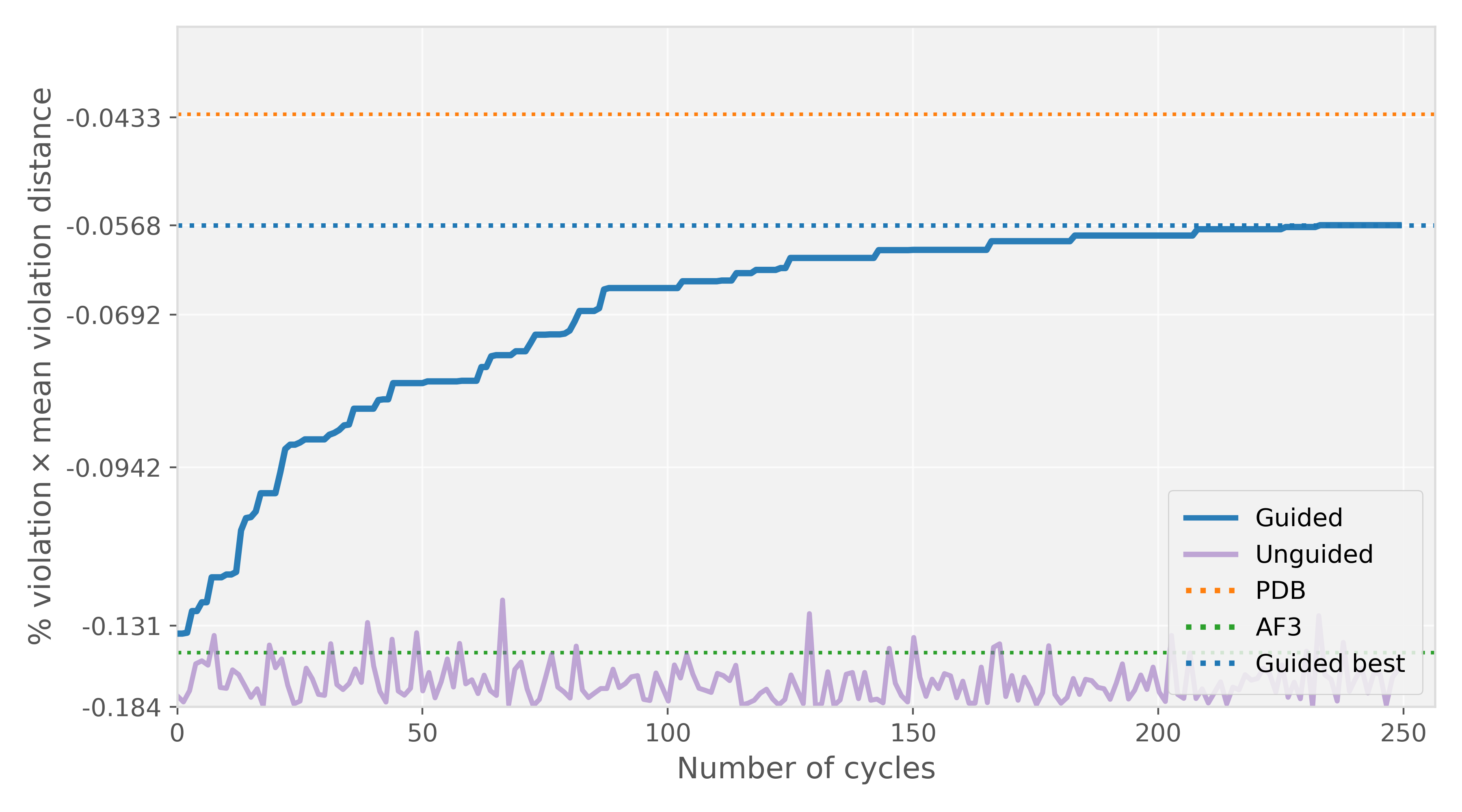}
     \caption{\textbf{1PQX NOE guidance.} 
    BioEmu guided with NOE-derived restraints. 
    Y-axis: fraction of violated restraints multiplied by mean violation distance. 
    Guidance yields consistent improvement in the best-structure metric.}
    \label{fig:noe_1pqx}
\end{figure}

This section provides extended results for NOE-guided experiments. 
In all cases, candidate structures were scored using the NOE violation metric 
(fraction of violated restraints multiplied by mean violation distance; see Appendix~\ref{appendix:technical_details}).  
Figures~\ref{fig:noe_1dec}–\ref{fig:noe_1pqx} report optimization traces for multiple proteins. Table \ref{tab:noe_metrics} reports metric comparisons in violation percentage, mean violation and median violation for the reference PDB, the AF3 prior and the best guided sample.
Consistently across all cases, BioEmu guided with NOE-derived restraints improves the best-structure metric over cycles, always outperforming the AlphaFold prior and the unguided BioEmu baseline, and often matching or surpassing the reference PDB performance.  
These consistent results demonstrate the effectiveness of the tailored genetic algorithm in leveraging experimental restraints to refine generative models toward experimentally consistent structures.

\begin{table}[H]
\centering
\begin{tabular}{llcccc}
\toprule
PDB & Method & Loss & Percent violation & Mean violation & Median violation \\
\midrule
\multirow{3}{*}{1dec}
  & PDB         & 0.1279 & 0.2211 & 0.5787 & 0.4981 \\
  & AF3         & 1.0658 & 0.3347 & 3.1846 & 1.5388 \\
  & Guided-best & \textbf{0.1184} & \textbf{0.2110} & \textbf{0.5613} & \textbf{0.3344} \\
\midrule
\multirow{3}{*}{2kkz}
  & PDB         & \textbf{0.1445} & 0.2629 & \textbf{0.5496} & 0.4081 \\
  & AF3         & 0.2836 & 0.3042 & 0.9324 & 0.5786 \\
  & Guided-best & 0.1492 & \textbf{0.2579} & 0.5784 & \textbf{0.4093} \\
\midrule
\multirow{3}{*}{1pqx}
  & PDB         & \textbf{0.0429} & \textbf{0.1215} & \textbf{0.3533} & \textbf{0.3389} \\
  & AF3         & 0.1409 & 0.1497 & 0.9417 & 0.3457 \\
  & Guided-best & 0.0594 & \textbf{0.1215} & 0.4894 & 0.3405 \\
\midrule
\multirow{3}{*}{2li3}
  & PDB         & 0.0259 & 0.0644 & 0.4016 & 0.3664 \\
  & AF3         & 1.1002 & 0.2441 & 4.5076 & 3.9301 \\
  & Guided-best & \textbf{0.0001} & \textbf{0.0068} & \textbf{0.0080} & \textbf{0.0076} \\
\midrule
\multirow{3}{*}{3bbg}
  & PDB         & 0.3200 & 0.5303 & \textbf{0.6035} & 0.4873 \\
  & AF3         & 0.7224 & 0.6086 & 1.1870 & 0.5474 \\
  & Guided-best & \textbf{0.3180} & \textbf{0.5126} & 0.6203 & \textbf{0.4672} \\
\bottomrule
\end{tabular}

\vspace{0.5em}
\caption{Comparison of NOE metrics across the AlphaFold (AF3) prior, the PDB reference, and the best structure in the guided population (min-loss). 
Bold indicates the best (lowest) value for each metric within a PDB. 
Details of the evaluation procedure are provided in Appendix~\ref{appendix:evaluation}.
}
\label{tab:noe_metrics}
\end{table}

\begin{figure}[H]
    \centering
    \includegraphics[width=0.7\linewidth]{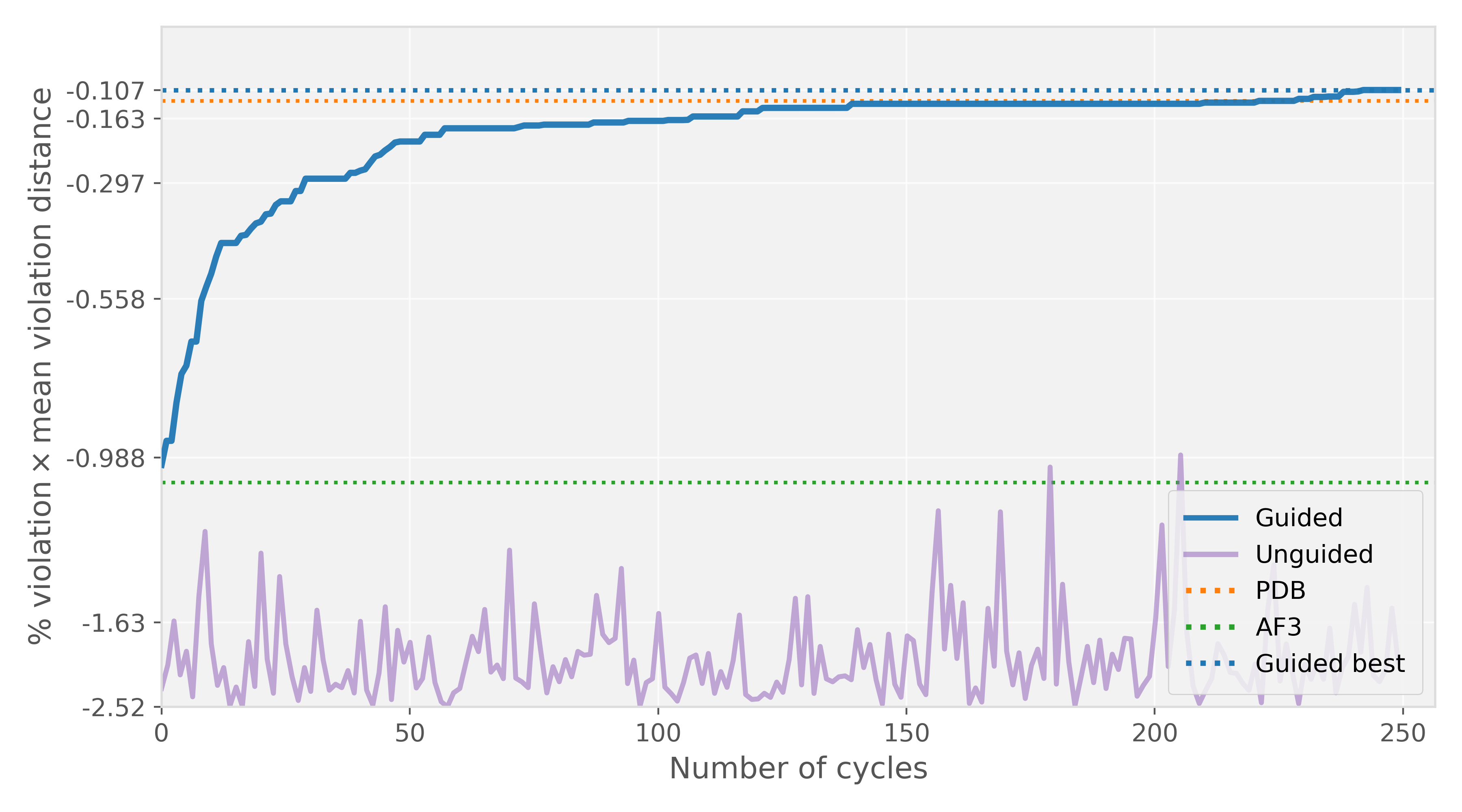}
     \caption{\textbf{1DEC NOE guidance.} 
    BioEmu guided with NOE-derived restraints. 
    Y-axis: fraction of violated restraints multiplied by mean violation distance. 
    Guidance yields consistent improvement in the best-structure metric, surpassing the PDB according to this measure.}
    \label{fig:noe_1dec}
\end{figure}

\subsubsection{NOE cumulative violation distributions}
\label{appendix:noe_violation_curves}

\begin{figure}[H]
    \centering
    \begin{subfigure}[t]{0.32\linewidth}
        \centering
        \includegraphics[width=\linewidth]{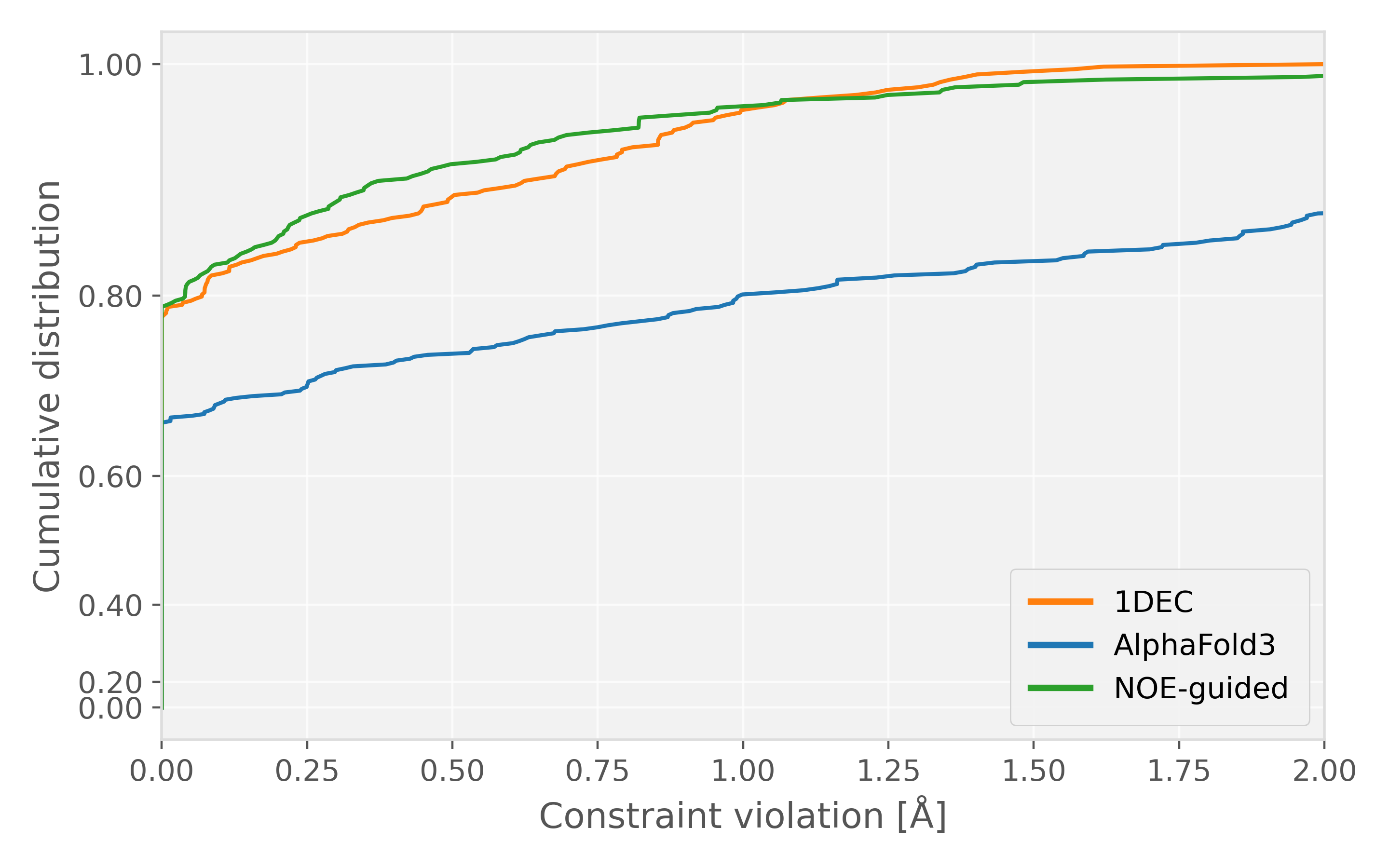}
        \caption{\texttt{1DEC}}
    \end{subfigure}\hfill
    \begin{subfigure}[t]{0.32\linewidth}
        \centering
        \includegraphics[width=\linewidth]{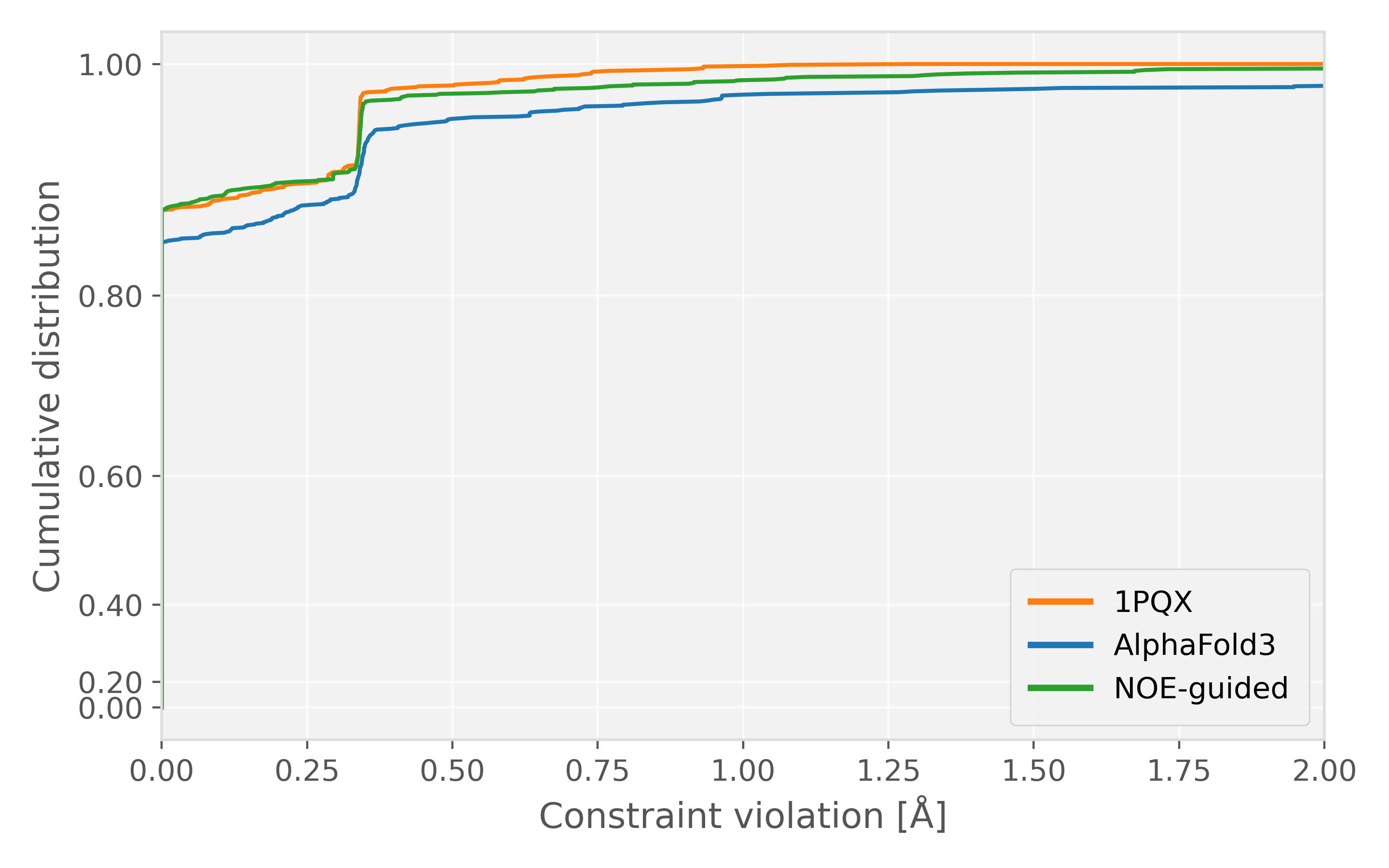}
        \caption{\texttt{1PQX}}
    \end{subfigure}\hfill
    \begin{subfigure}[t]{0.32\linewidth}
        \centering
        \includegraphics[width=\linewidth]{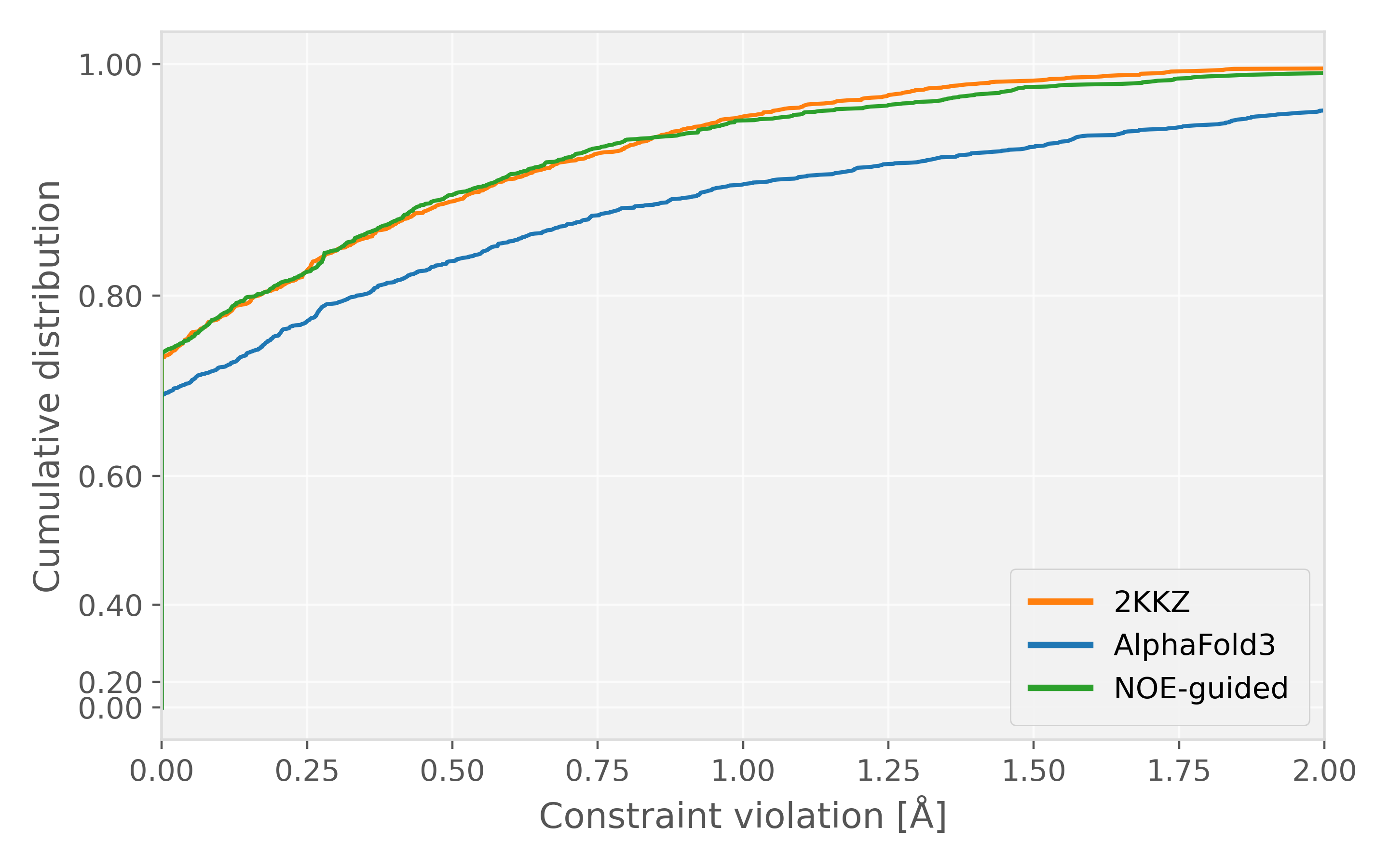}
        \caption{\texttt{2KKZ}}
    \end{subfigure}

    \vspace{0.6em}

    \begin{minipage}{0.66\linewidth}
      \centering
      \begin{subfigure}[t]{0.48\linewidth}
          \centering
          \includegraphics[width=\linewidth]{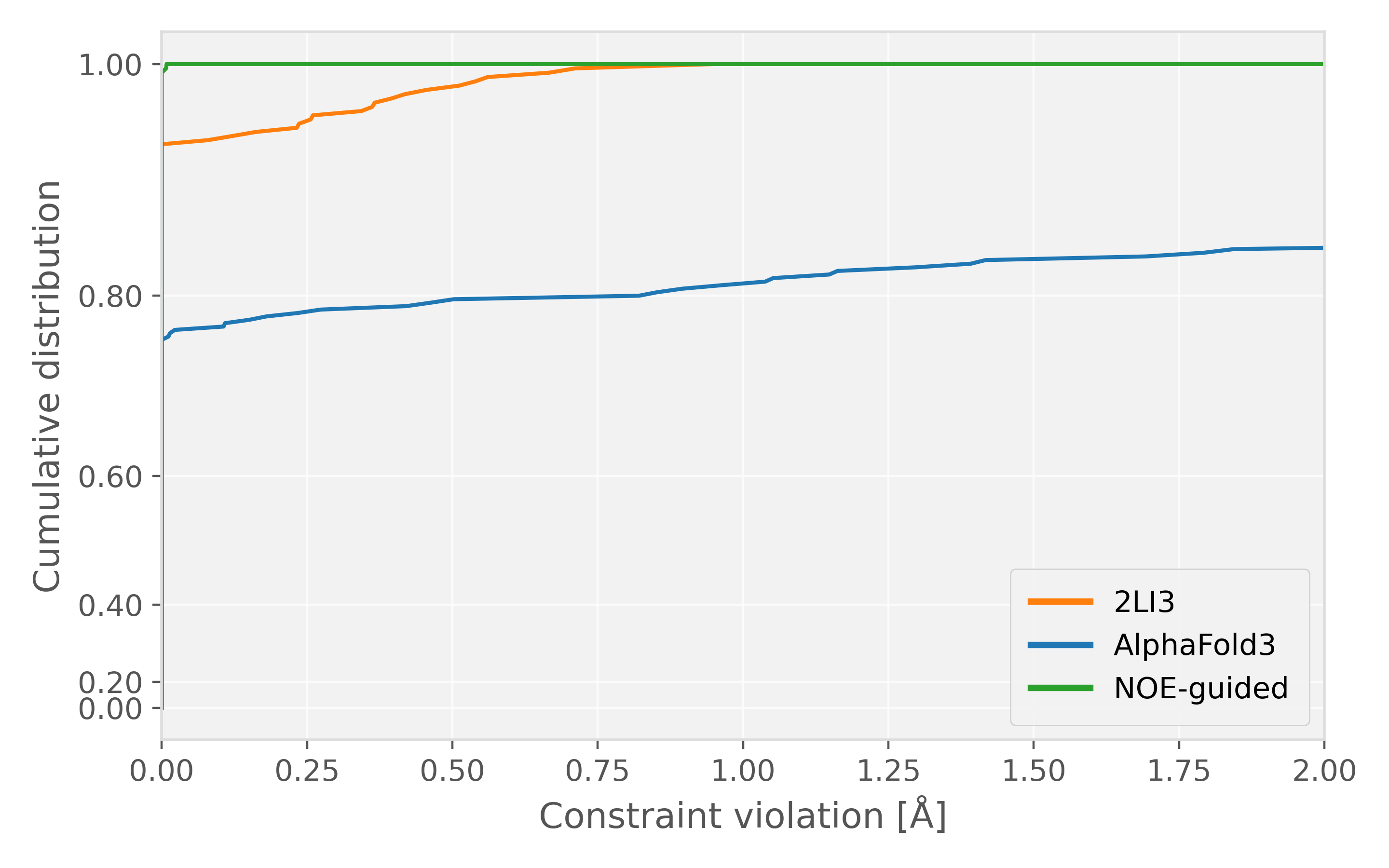}
          \caption{\texttt{2LI3}}
      \end{subfigure}\hfill
      \begin{subfigure}[t]{0.48\linewidth}
          \centering
          \includegraphics[width=\linewidth]{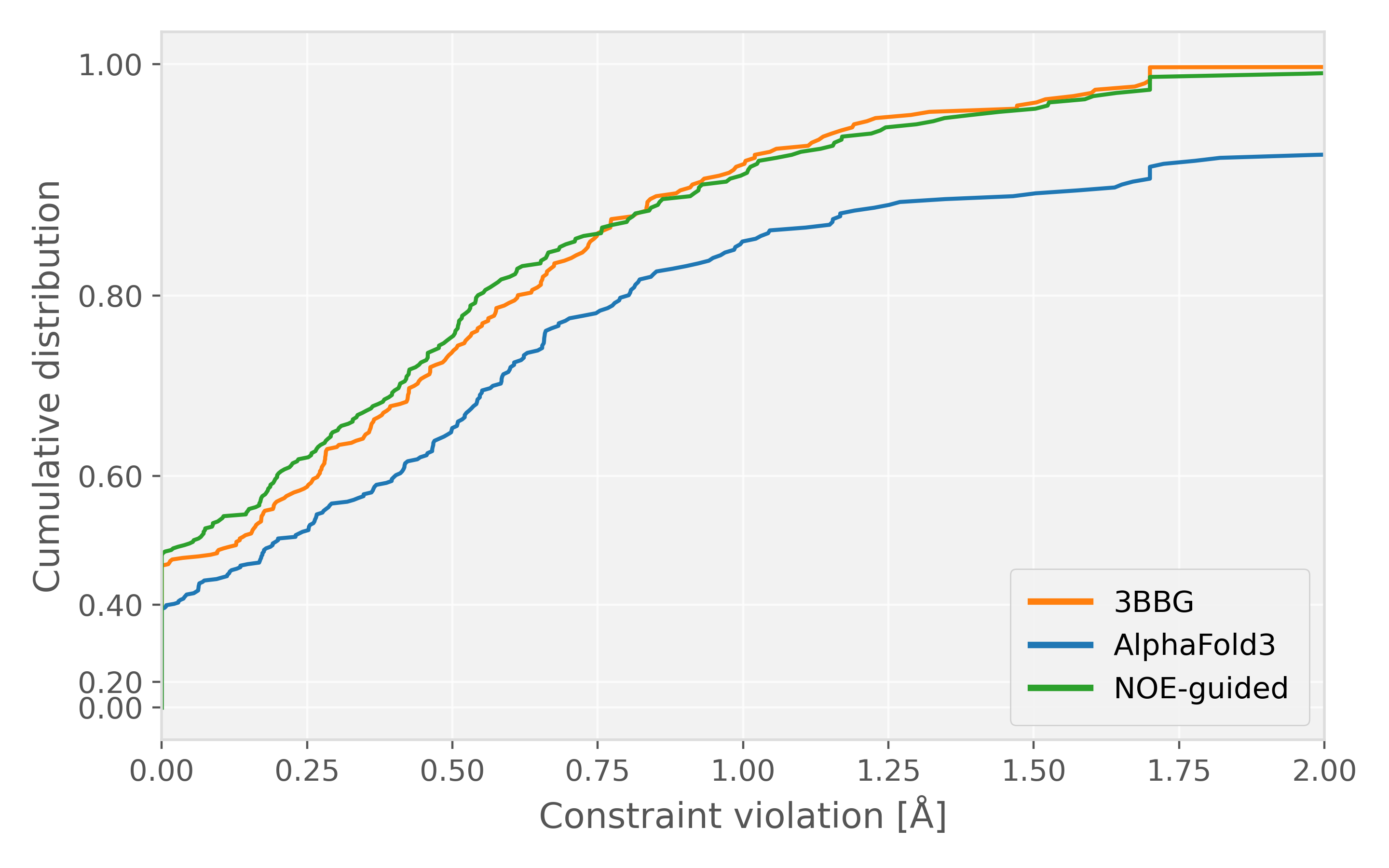}
          \caption{\texttt{3BBG}}
      \end{subfigure}
    \end{minipage}

    \caption{\textbf{Cumulative violation distributions for NOE restraints.} Faster-rising curves indicate fewer/smaller violations. Guided consistently outperforms AF3 and is comparable to or better than the PDB across targets.}
    \label{fig:noe_violation_ecdfs}
\end{figure}

\noindent\textbf{Setup.} For each target we plot the cumulative distribution of NOE violation magnitudes (including zeros) in Fig.~\ref{fig:noe_violation_ecdfs}, i.e., the fraction of restraints with violation $\leq x$ as a function of the threshold $x$ (in \AA). Curves that rise faster/earlier indicate fewer and/or smaller violations. Each panel compares the reference PDB structure, the AlphaFold3 (AF3) prior, and the best guided structure (minimum NOE loss) from our search.

\noindent\textbf{Summary.} Across all targets, the guided curve consistently dominates the AF3 curve (higher at nearly all thresholds), showing uniformly fewer/smaller violations than the AF3 prior. Relative to the PDB reference, guided is either comparable or better over most of the range (notably strong on \texttt{2LI3}), and on the remaining cases it closely tracks the PDB curve. These results mirror the scalar NOE metrics in the main text/table and visualize how guidance reduces both violation frequency and severity.

\subsection{Chemical shift experiments}
\label{appendix:additional_cs}

This section provides extended results for chemical shift–guided experiments. 
Structures guided with UCBShift-predicted shifts are shown in Fig.~\ref{fig:ucbshift_struct}.  
The chemical shift metric, defined in Appendix~\ref{appendix:cs_model}, consistently improved over optimization cycles, although plateauing behavior was observed at certain stages.  
Despite this metric improvement, the resulting guided structures did not converge toward the experimental PDB ensembles, underscoring the discrepancy between chemical shift agreement and structural similarity.  
These results highlight both the utility of chemical shift data for guidance and the current limitations of predictors such as UCBShift, which lack explicit physical grounding.

\begin{figure}[H]
    \centering
    \begin{subfigure}{0.7\textwidth}
        \centering
        \includegraphics[width=\linewidth]{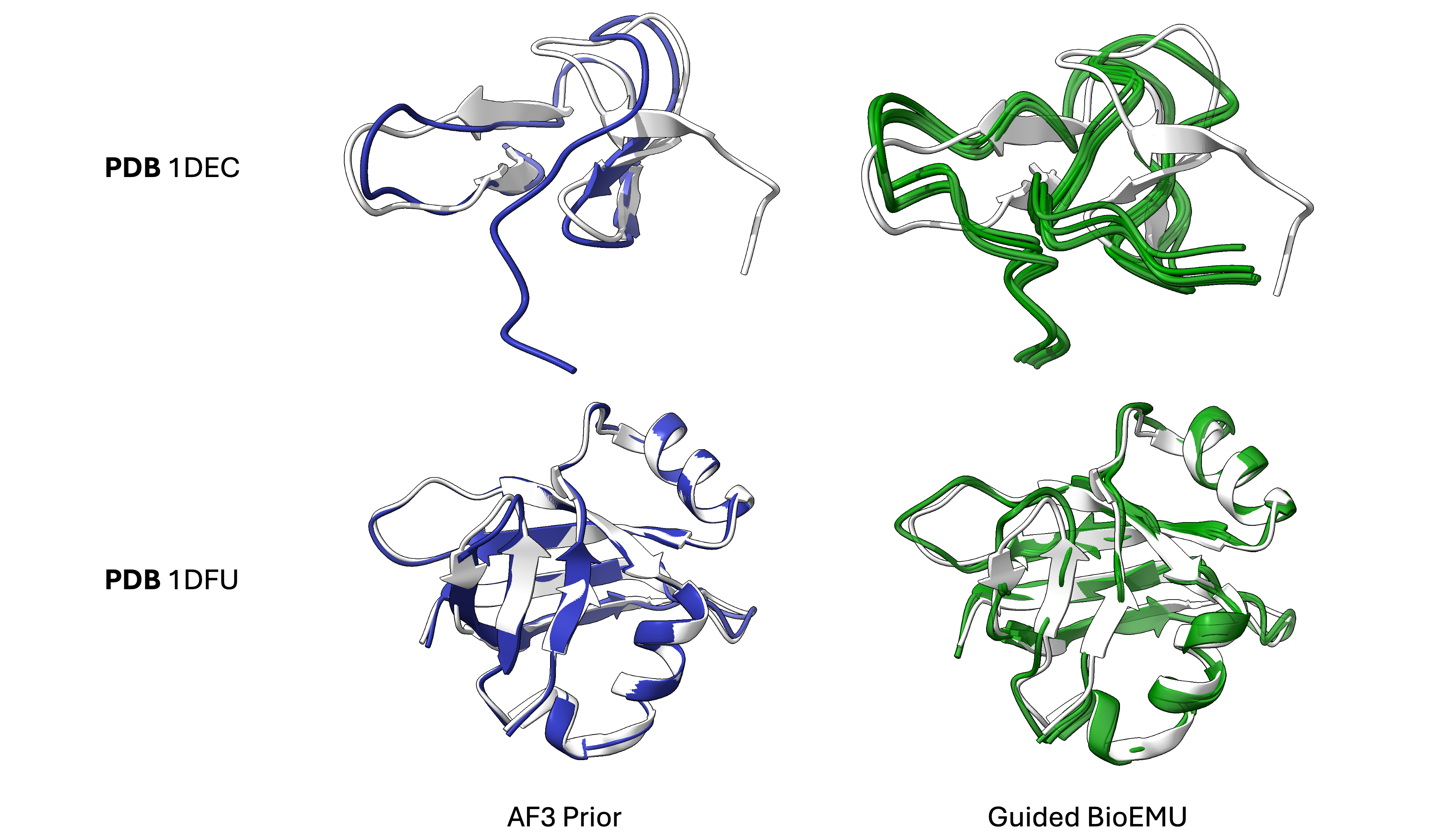}
    \end{subfigure}
    \caption{\textbf{Chemical shift guided structures.} 
    Guiding BioEmu with UCBShift shifts predictions. Reference PDB structures shown in shite. 
    1DEC was guided with synthetic PDB-derived shifts. 1DFU was guided with experimental shifts (BMRB 4395).
    }
    \label{fig:ucbshift_struct}
\end{figure}

\section{Plateau in Metric Graphs}\label{appendix:plateau}

When running our genetic-guided algorithm with a constant perturbation time, we observed plateaus in the evaluation metric.  
To address this, we modified the perturbation diffusion time to follow a linear schedule.  
This adjustment helps because, as the search progresses, the required mutations need to become more fine-grained.  
A fixed diffusion time cannot adequately support convergence, since the resulting mutations introduce excessive structural changes. In contrast, gradually reducing the diffusion time across cycles enables faster and more accurate convergence.

However, plateaus still occur.  
We hypothesize that this happens because the genetic search requires spending varying numbers of cycles at different perturbation times, and a simple linear schedule cannot capture this behavior.  
As future work, we plan to design a more adaptive scheduling strategy in which the perturbation time is adjusted dynamically—only when a plateau is detected.  

This trend is illustrated in Figures~\ref{fig:noe_1dec}–\ref{fig:noe_1pqx}.

\section{Technical details}\label{appendix:technical_details}

The following section provides full technical details of our method.  
We first describe the \textbf{score functions} used to evaluate candidate structures during genetic optimization (Section~\ref{appendix:score_models}), including pairwise distance matching, NOE-derived distance restraints, and chemical shift prediction with UCBShift.  
Next, we outline the \textbf{evaluation metrics} (Section~\ref{appendix:evaluation}) used to quantify structural agreement with experimental data.  
We then provide details of \textbf{data curation} (Section~\ref{appendix:data_curation}), specifying how reference structures and experimental constraints were selected and processed for each modality.  
Finally, we summarize \textbf{runtime analysis} (Section~\ref{appendix:runtime}) and \textbf{hyperparameter settings} (Section~\ref{appendix:hyperparameters}) to support replication of our experiments.

\subsection{Score models}
\label{appendix:score_models}

To evaluate candidate protein structures in our genetic algorithm, we used three complementary scoring models: NOE-based distance restraints, generic pairwise distance matching, and chemical shift prediction (UCBShift~\citep{ptaszek2024ucbshift2}). 

\subsubsection{Pairwise distance model}\label{appendix:pairwise_model}
The pairwise model provides a general framework for comparing inter-atomic distances against reference values.  
Inputs are a CSV file of atom pairs with target distances. Given a structure, distances are computed as
\[
d_{ij} = \| \mathbf{r}_i - \mathbf{r}_j \|_2.
\]
Three loss modes are supported:
\begin{itemize}
    \item \texttt{l2}: mean squared error between predicted and target distances,
    \item \texttt{l1}: mean absolute error,
    \item \texttt{topk}: mean absolute error of the $k$ largest deviations, highlighting worst-case mismatches.
\end{itemize}
In practice, we found that the \texttt{l1} formulation provided the most stable and informative signal for guiding the genetic algorithm, and it was therefore used as the default.

\subsubsection{NOE restraint model}\label{appendix:noe_model}
The NOE scoring function enforces satisfaction of nuclear Overhauser effect (NOE)–derived distance restraints. Input restraints are provided as a CSV file with atom–atom pairs, residue indices, and upper/lower distance bounds.  

Since BioEmu does not generate explicit hydrogen atoms, we approximate NOE restraints by enforcing them on the heavy atoms to which the hydrogens are covalently bound. For example, H–H restraints are mapped to the corresponding C–C or C–N distances. This approximation preserves the geometric character of the NOE but avoids inconsistencies due to missing hydrogens. 

For each experimental restraint between atoms $i$ and $j$, we compute the interatomic distance
\[
d_{ij} = \| \mathbf{r}_i - \mathbf{r}_j \|_2,
\]
and measure its deviation from the allowed bounds $\ell_{ij}$ and $u_{ij}$ as
\[
L_{ij} = \max\!\big(0, \, \ell_{ij} - d_{ij}\big) + \max\!\big(0, \, d_{ij} - u_{ij}\big).
\]
Thus $L_{ij}=0$ whenever the restraint is satisfied, and grows linearly with the degree of violation otherwise.  

Because NMR assignments are often uncertain, restraints are grouped into \emph{ambiguous sets} $G \subseteq \mathcal{R}$ with an effective \texttt{OR}-logic: it is sufficient that one member of the set is satisfied.  
Accordingly, the violation of a group is defined as the smallest of its constituent penalties,
\[
L_G = \min_{(i,j)\in G} L_{ij}.
\]

Let $\mathcal{G}$ be the collection of all such groups, and $\mathcal{V} = \{ G \in \mathcal{G} : L_G > 0 \}$ the subset of violated groups.  
The overall NOE score combines both frequency and severity of violations:
\[
L_{\mathrm{NOE}} = 
\frac{|\mathcal{V}|}{|\mathcal{G}|} \cdot 
\frac{1}{|\mathcal{V}|} \sum_{G \in \mathcal{V}} L_G.
\]
In other words, the final loss equals the fraction of violated groups multiplied by the average magnitude of those violations.  

This design penalizes both widespread mild inconsistencies and a few severe outliers.  
Since it involves $\max$ and $\min$ operations together with discrete violation indicators, the function is \emph{non-differentiable}.  
We therefore use $L_{\mathrm{NOE}}$ strictly as a black-box objective in our genetic search.

\paragraph{Distances vs. peak intensities.}  
NOE signals in NMR originate from dipolar couplings between protons within roughly $6$~\r{A}.  
The intensity of a cross-peak in a NOESY spectrum follows an inverse-sixth power dependence on the internuclear distance, $I \propto r^{-6}$.  
After peak assignment, these intensities are typically converted into effective distance restraints according to
\[
r = r_{\text{ref}} \left(\frac{I_{\text{ref}}}{I}\right)^{1/6},
\]
where $r_{\text{ref}}$ and $I_{\text{ref}}$ denote the distance and intensity of a reference pair.  

For structural ensembles, physical consistency requires computing the mean of the calibrated intensities across conformations and then transforming this average intensity back into a distance.  
Directly averaging distances neglects the $r^{-6}$ relationship and can bias the restraints.  
In the present work we use the simpler distance-based formulation, while a fully intensity-averaged treatment will be incorporated in future extensions.

\paragraph{Ensemble effects.}  
Because NOE cross-peaks reflect population-weighted averages over conformational ensembles, restraints are fundamentally ensemble quantities.  
In this work, however, we applied the constraints independently to each generated structure. 
A more rigorous treatment would evaluate NOE violations over an ensemble of conformations, thereby capturing conformational heterogeneity; we leave this extension for future work.

\subsubsection{Chemical shift model (UCBShift)}\label{appendix:cs_model}
To incorporate chemical shift data, we used UCBShift~\citep{ptaszek2024ucbshift2}, a widely used chemical shift predictor trained on large experimental databases.  
Our implementation calls both the \texttt{X} and \texttt{Y} modules of UCBShift to obtain predictions for the supplied structure via the \texttt{calc\_sing\_pdb} routine.  
Experimental reference shifts are read from a CSV file and aligned by residue index to the predicted values.  

The loss is computed as either
\[
L_{\mathrm{l1}} = \frac{1}{N} \sum_i | y_i - \hat{y}_i |, 
\quad
L_{\mathrm{l2}} = \frac{1}{N} \sum_i ( y_i - \hat{y}_i )^2,
\]
depending on the chosen mode.  

Although the $L_1$ and $L_2$ loss functions are differentiable, the overall score is \emph{not differentiable}, because the UCBShift algorithm itself is non-differentiable.
Thus, gradients with respect to structural coordinates are not available, and the model can only be used as a black-box scoring function within our genetic algorithm.

\subsection{Evaluation metric}\label{appendix:evaluation}

\subsubsection{Pairwise distances}

Evaluation for the pairwise distance experiments is based directly on the loss functions defined in Appendix~\ref{appendix:pairwise_model}.  
Given reference distances and predicted distances, the score is computed using the chosen formulation (\texttt{l1}, \texttt{l2}, or \texttt{topk}), with the $L_1$ version serving as our default.  
Thus, the pairwise evaluation metric coincides exactly with the pairwise loss used during optimization.

\subsubsection{NOE restraints}

We evaluate structural agreement with NOE restraints using the violation terms defined in Appendix~\ref{appendix:noe_model}.  
For each restraint group $G \in \mathcal{G}$ we obtain a group-level violation $L_G$, where $L_G = 0$ indicates satisfaction and $L_G > 0$ indicates a violation.  
From these values we report three descriptive metrics:
\begin{itemize}
    \item \textbf{Violated NOE restraints percentage:} the fraction of groups with nonzero violation,  
    \[
    p_{\mathrm{viol}} = \frac{|\{ G \in \mathcal{G} : L_G > 0 \}|}{|\mathcal{G}|}.
    \]

    \item \textbf{Median violation:} the median magnitude of violations across the violated groups (reported in \AA),  
    \[
    m_{\mathrm{viol}} = \mathrm{median}\big(\{ L_G : L_G > 0 \}\big).
    \]

    \item \textbf{Mean violation:} the mean magnitude of violations across the violated groups (reported in \AA),  
    \[
    \mu_{\mathrm{viol}} = \frac{1}{|\{ G \in \mathcal{G} : L_G > 0 \}|} \sum_{G : L_G > 0} L_G.
    \]
\end{itemize}

As in the NOE model, we also summarize performance with a scalar loss
\[
L_{\mathrm{NOE}} = p_{\mathrm{viol}} \cdot \mu_{\mathrm{viol}},
\]
which penalizes both the prevalence and severity of violations.

\subsubsection{Chemical shift}

Evaluation against chemical shift data follows the loss definitions given in Appendix~\ref{appendix:cs_model}.  
Predicted shifts $\hat{y}_i$ are compared to experimental reference values $y_i$ using either the $L_1$ or $L_2$ loss, with $L_1$ as the default.  
Accordingly, the chemical shift evaluation metric coincides with the scoring function used during optimization.

\subsection{Data curation}\label{appendix:data_curation}

\subsubsection{Pairwise distances}

For the pairwise distance experiments we selected the protein 4OLE, which corresponds to a domain from the NBR1 gene, a neighbor of human BRCA1 that has been implicated in breast cancer.  
This entry is particularly challenging because it contains two alternative conformations (altloc A and altloc B).  
Conformation A adopts a helical arrangement, whereas conformation B forms a strand.  
Since BioEmu by default generates structures closer to altloc B, we instead used altloc A as the reference.  
The main structural differences are concentrated in residues \text{60--68}, alongside smaller variations that are less consequential.  
From the reference structure we computed all C$_\alpha$–C$_\alpha$ pairwise distances and retained only those pairs within 5\,\text{\AA}.  
This cutoff was chosen to approximate the short-range nature of NOE-derived contacts, providing a simplified proxy for NMR-like restraints.  
The resulting set of pairwise contacts was used as the experimental input for the genetic guidance procedure.

\subsubsection{NOE restraints}
\paragraph{Structure selection}  
We evaluated our method on proteins where the BioEmu generator alone produced inaccurate folds.  
From the benchmark of \citep{mcdonald2023benchmarking}, we first considered the $20$ peptides reported as most challenging for AlphaFold.  
This yielded three peptides—\texttt{1DEC}, \texttt{2LI3}, and \texttt{3BBG} —on which BioEmu also generated erroneous or partially misfolded conformations.  

We further selected proteins from the 100 NMR spectra database~\cite{klukowski2024100}, choosing \texttt{1PQX} and \texttt{2KKZ} because BioEmu struggled to reproduce their experimental folds.  
Together, these five cases form a targeted testbed: experimentally validated NMR ensembles that expose weaknesses of both AlphaFold and BioEmu, allowing us to test whether our genetic guidance procedure can repair BioEmu’s predictions.

\paragraph{NOE restraint extraction}  
NOE-derived distance restraints were obtained directly from the NMR STAR files associated with each PDB entry.  
We parsed the restraint tables using the \texttt{pynmrstar} library, selecting only those entries explicitly labeled as NOE constraints (\texttt{\_Gen\_dist\_constraint\_list.Constraint\_type}).  
Other types of structural information present in the files—such as hydrogen-bond, dihedral, or RDC-derived restraints—were excluded so that the scoring relied solely on NOE distance geometry.  

Ambiguous assignments involving multiple possible proton pairs were retained without disambiguation, in order to reflect the inherent uncertainty of NMR peak assignment.  
Lower and upper bounds were taken directly from the STAR fields \texttt{\_Gen\_dist\_constraint\_list.Distance\_lower\_bound\_val} and \texttt{\_Gen\_dist\_constraint\_list.Distance\_upper\_bound\_val}, respectively.  
When a lower bound was missing, it was explicitly set to 0\,\text{\AA} to enforce a physically valid minimum distance.  
All bounds were kept in their original \text{\AA}ngström units, with no additional normalization or thresholding, to preserve fidelity to the experimental input data.

\subsubsection{Chemical shifts}


The evaluation of a realistic case was done using experimental chemical shifts from RefDB~\citep{zhang2003refdb}, a curated subset of the BMRB~\citep{hoch2023biological} containing calibrated assignments.  


Additional details are described in Section.~\ref{sec:chemical_shift}

\subsection{Runtime analysis}
\label{appendix:runtime}

Unless otherwise stated, all runs were executed on a single NVIDIA H100 GPU with 30 CPU cores used for parallel score evaluation. Excluding the UCBShift experiments, wall-clock time varied between 1.0 and 3.5 hours per target, primarily as a function of sequence length and atom count. Small proteins such as \texttt{1DEC} completed in roughly 1 hour, whereas larger targets like \texttt{4OLE} required about 3 hours. The UCBShift runs exhibit higher variance due to predictor latency and the larger number of generations (Section~\ref{appendix:hyperparameters}); we therefore report only cycle counts for those experiments.

\vspace{0.5em}
\noindent\textit{Hardware:} 1$\times$H100 GPU, 30 CPU cores; parallelized black-box scoring.

\subsection{Hyperparameters}
\label{appendix:hyperparameters}

We tuned four hyperparameter families: (i) the number of generations (cycles), (ii) the diffusion-time scheduler that maps generation index $k$ to a perturbation time $t_k$, (iii) the tournament size for parent selection, and (iv) the elitism keep for survivor selection.

\paragraph{Population and selection.}
Unless noted otherwise, we used a population size of 100, tournament size = 5, and elitism keep = 5. These values provided a stable exploration–exploitation balance across targets.

\paragraph{Cycles (generations).}
Most experiments ran for 250 cycles. Exceptions are summarized below:
\begin{center}
\begin{tabular}{@{}lcc@{}}
\toprule
Target & Modality & Cycles \\
\midrule
\texttt{1DEC} & UCBShift (chemical shifts) & \textbf{1700} \\
\texttt{1DFU} & UCBShift (chemical shifts) & \textbf{600} \\
\texttt{2LI3} & NOE & \textbf{153} \\
\textit{All others} & Pairwise / NOE & \textbf{250} \\
\bottomrule
\end{tabular}
\end{center}

\paragraph{Diffusion-time scheduler.}
The scheduler specifies which perturbation diffusion time $t_k$ is used at generation $k$. We tuned this mapping to control the global$\rightarrow$local search progression. Concretely, we select $(t_{\min}, t_{\max})$ and a schedule shape, then set $t_k$ accordingly. In practice, early generations use larger $t_k$ to promote broader moves; later generations reduce $t_k$ to focus on local refinements. For difficult objectives (e.g., UCBShift) we biased the schedule toward spending more generations at intermediate $t_k$ to mitigate early plateaus. For all other experiments, we started from diffusion index 0 and increased it by 1 every 4 cycles, capping the maximum index at 40. The generative process spans indices 0 (completely noisy) through 50 (completely clean).


\end{document}